\documentclass{paper}
\usepackage[utf8]{inputenc}
\usepackage{graphicx}
\usepackage{caption}
\usepackage{subcaption}
\usepackage{amsfonts}
\usepackage{mathtools}

\usepackage[accepted]{arxiv}
\usepackage[switch]{lineno}

\begin{document}

\twocolumn[
\icmltitle{Interpretable Joint Event-Particle Reconstruction for Neutrino Physics at NOvA with Sparse CNNs and Transformers}



\icmlsetsymbol{equal}{*}

\begin{icmlauthorlist}
\icmlauthor{Alexander Shmakov}{equal,uci-cs}
\icmlauthor{Alejandro Yankelevich}{equal,uci-ps}
\icmlauthor{Jianming Bian}{uci-ps}
\icmlauthor{Pierre Baldi}{uci-cs}
\icmlauthor{for the NOvA Collaboration}{}

\end{icmlauthorlist}

\icmlaffiliation{uci-cs}{School of Information and Computer Sciences, University of California Irvine, USA}
\icmlaffiliation{uci-ps}{Department of Physics and Astronomy, University of California Irvine, USA}

\icmlcorrespondingauthor{Alexander Shmakov}{ashmakov@uci.edu}

\icmlkeywords{Attention, Physics, Interpretability}

\vskip 0.3in
]

\printAffiliationsAndNotice{\icmlEqualContribution} 

\begin{abstract}
    The complex events observed at the NOvA long-baseline neutrino oscillation experiment contain vital information for understanding the most elusive particles in the standard model. The NOvA detectors observe interactions of neutrinos from the NuMI beam at Fermilab. Associating the particles produced in these interaction events to their source particles, a process known as reconstruction, is critical for accurately measuring key parameters of the standard model. Events may contain several particles, each producing sparse high-dimensional spatial observations, and current methods are limited to evaluating individual particles. To accurately label these numerous, high-dimensional observations, we present a novel neural network architecture that combines the spatial learning enabled by convolutions with the contextual learning enabled by attention. This joint approach, TransformerCVN, simultaneously classifies each event and reconstructs every individual particle's identity. TransformerCVN classifies events with 90\% accuracy and improves the reconstruction of individual particles by 6\% over baseline methods which lack the integrated architecture of TransformerCVN. In addition, this architecture enables us to perform several interpretability studies which provide insights into the network's predictions and show that TransformerCVN discovers several fundamental principles that stem from the standard model.
\end{abstract}

\section{Introduction}
With the increasingly widespread use of machine learning in the physical sciences, the lack of interpretability of deep neural networks trained is a common concern. The rapid increase in data collected in high energy physics has lead to the importance of machine learning methods for extracting meaningful results. Developing interpretable neural networks for these scientific applications is vital for building a more fundamental understanding of this data. A common task in particle physics which requires such data is \textit{reconstruction}, which is concerned with mapping a low-level, high-dimensional observation to more fundamental objects in the standard model. This reconstruction is key to understanding the observations produced by particle colliders since it allows us to compare our theoretical understanding of the universe to phenomenon observed within a detector.
    
\subsection{Event Reconstruction at NOvA}
\begin{figure}
    \centering
    \includegraphics[width=0.5\textwidth]{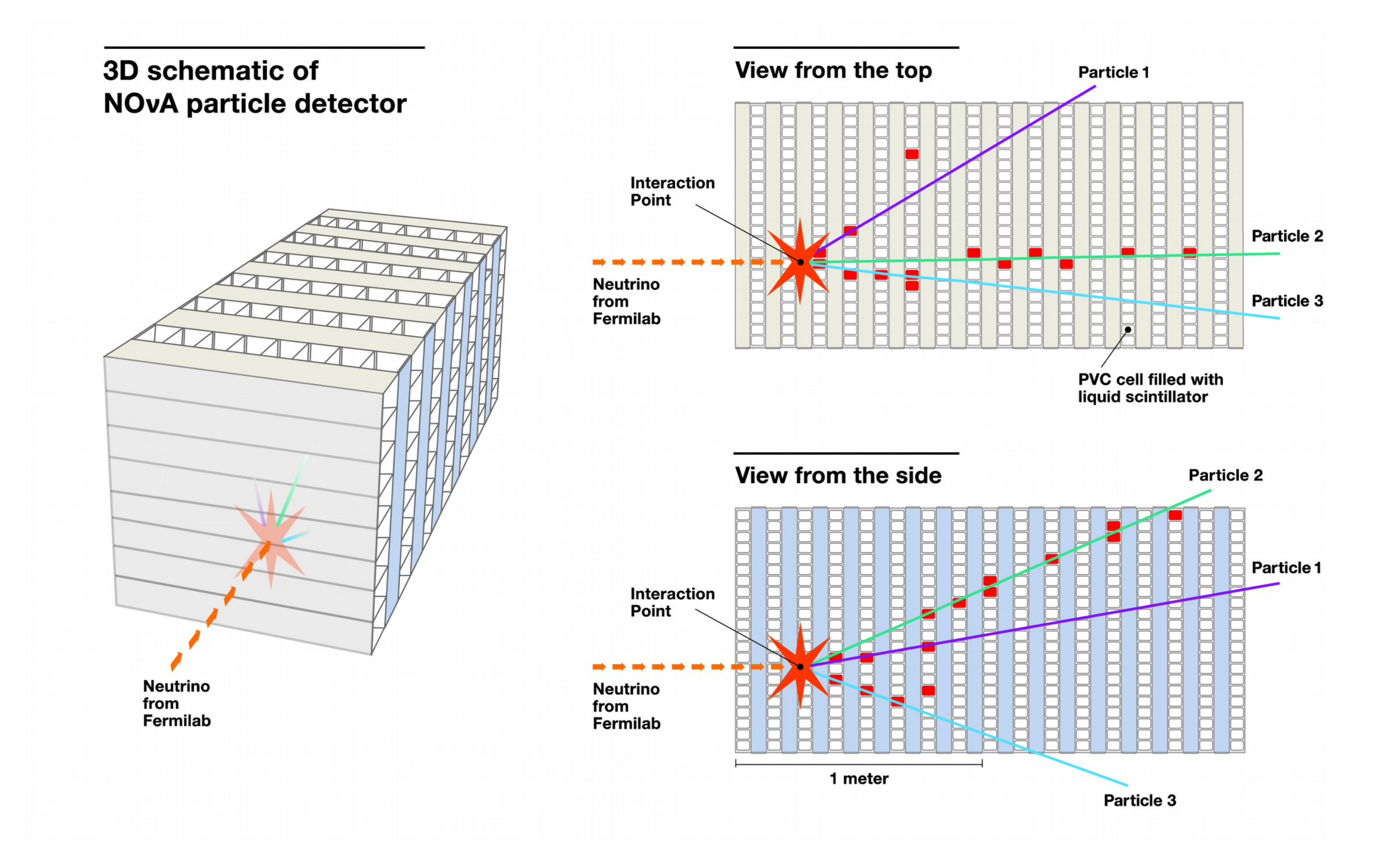}
    \caption{Schematic of NOvA detector and generation of top and side views from vertical and horizontal planes respectively.}
    \label{fig:schematic}
    \vspace{-12pt}
\end{figure}

NOvA is a long-baseline neutrino oscillation experiment using Fermilab’s NuMI beam. The experiment consists of two functionally identical detectors 809km apart formed from plastic extrusions filled with a liquid scintillator. The far detector consists of 896 alternating horizontal and vertical planes, and each plane contains 384 4 cm x 6 cm x 155 cm cells for a total of 344,064 cells \cite{nova_technical}. By gathering the vertical planes to generate an XZ view and gathering the horizontal planes to generate a YZ view, 3D reconstruction of particle tracks is possible (Figure \ref{fig:schematic}).

\begin{figure}
    \centering
    \includegraphics[width=0.5\textwidth]{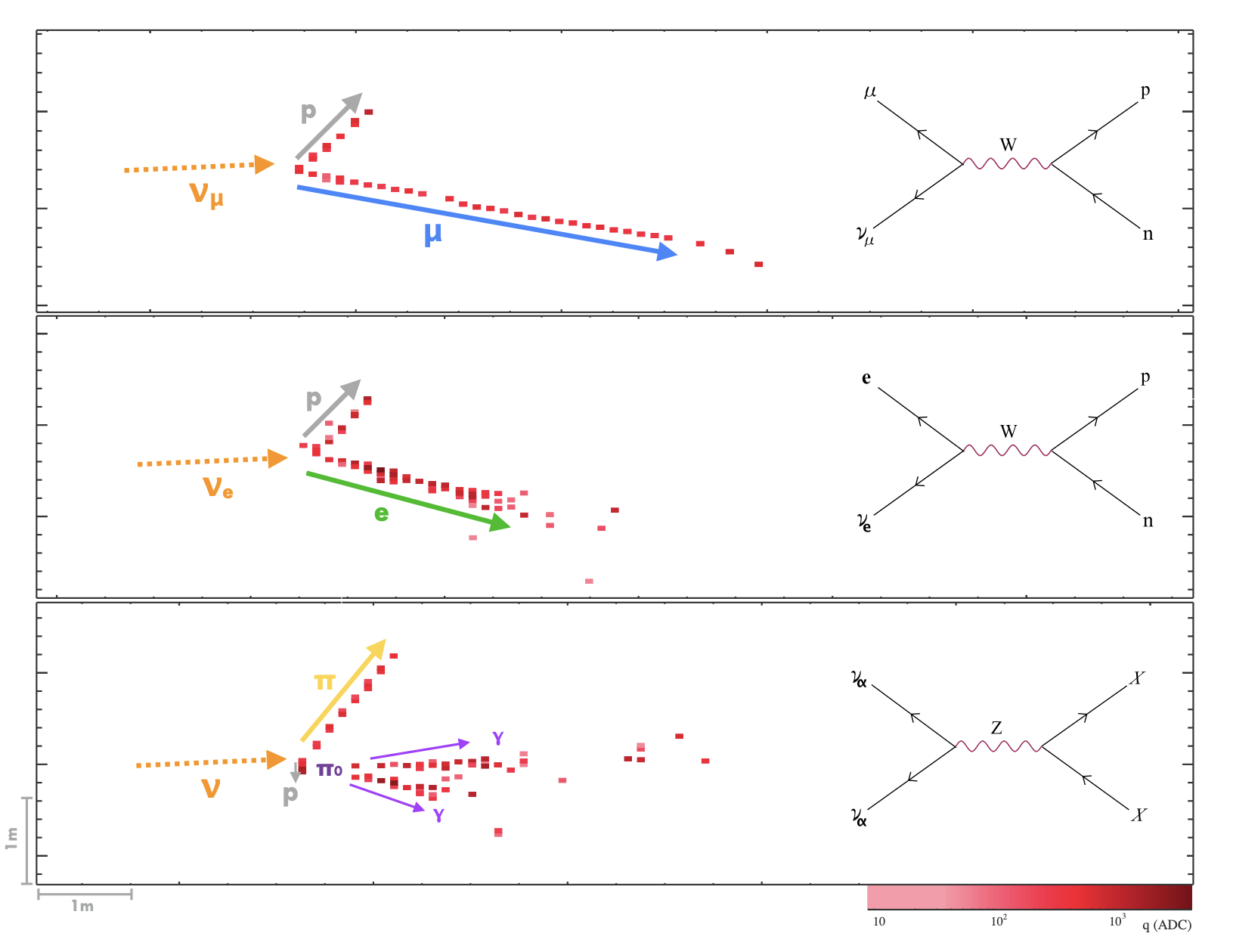}
    \caption{Typical far detector event displays of $\nu_\mu$ CC (top), $\nu_e$ CC (middle), and NC (bottom) neutrino interactions.}
    \label{fig:interactions}
\end{figure}

The NuMI beam is predominantly composed of muon neutrinos, $\nu_\mu$. The collaboration studies neutrino oscillations by detecting the rate of disappearance of $\nu_\mu$ and appearance of electron neutrinos, $\nu_e$, at the NOvA far detector through charged current (CC) interactions. In a CC interaction, the neutrino interacts with a nucleon and produces a lepton and a hadronic component. A $\nu_\mu$ CC interaction will produce a $\mu$ that, as a minimum ionizing particle, typically leaves a long track whereas a $\nu_e$ CC interaction will produce an $e$ that leaves a shower-like pattern of cell hits. Neutrinos can also interact via the neutral current (NC) interactions, but these interactions only produce a hadronic component, so the neutrino flavor cannot be determined (Figure \ref{fig:interactions}).

\begin{figure*}[t]
    \centering
    \includegraphics[width=0.8\textwidth]{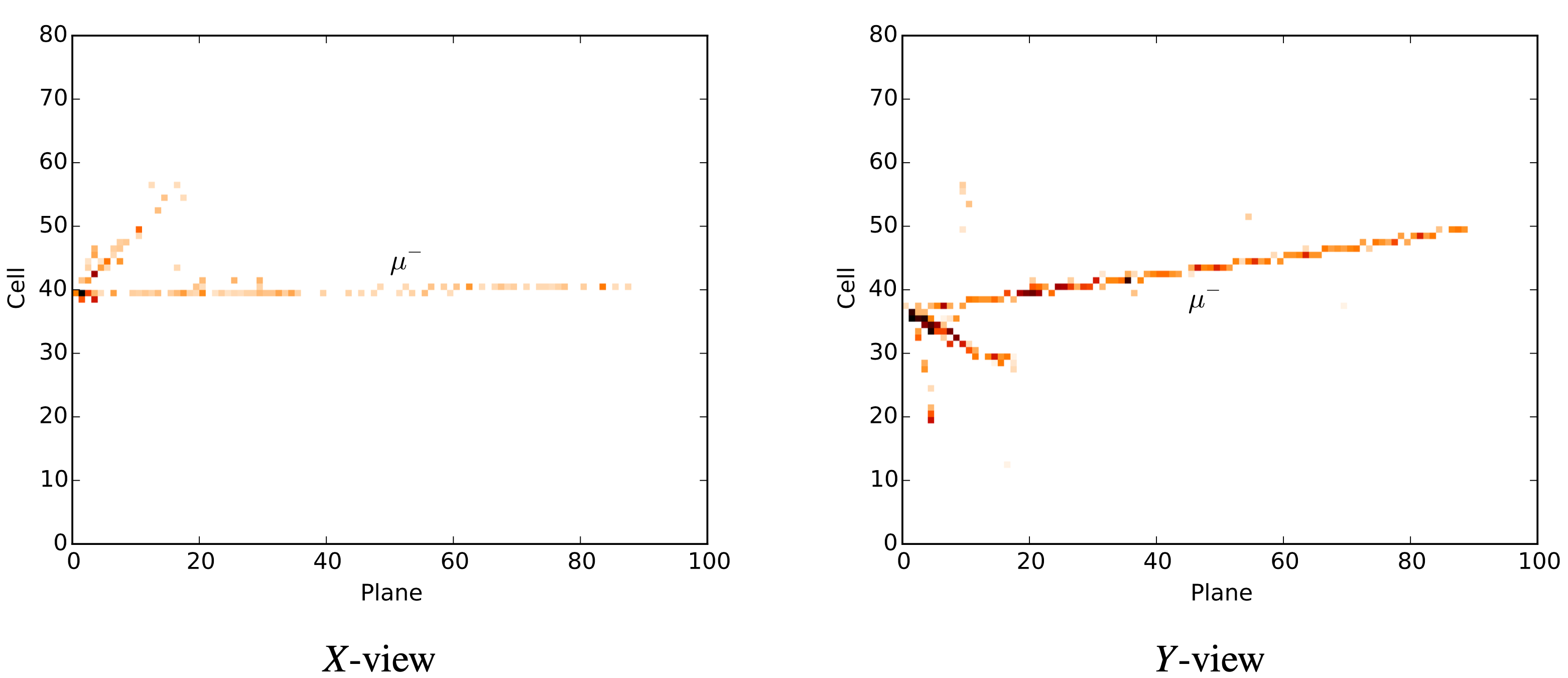}
    \caption{An Example pixel-map for a $\nu_\mu$ CC event.}
    \label{fig:pixel_maps}
\end{figure*}

The rectangular prism shape of the detector and cells naturally creates event display images in two views. NOvA's previous oscillation analysis \cite{oscillation} makes use of a CNN refered to as Convolution Visual Network (CVN) \cite{cvn}. The CVN trained for event classification, EventCVN, takes as input two 100 x 80 images cropped around the reconstructed event, which we refer to as \textit{pixel-maps}. To generate the pixel-maps, potential neutrino interaction events are split into clusters of energy deposits called "slices." The pixel-maps are cropped such that the first cell hit along the beam direction is placed in the first column along the z-axis and the hits are centered along the x or y direction as in Figure \ref{fig:pixel_maps}. The size of these pixel-maps corresponding to a 14.52 m x 4.18 m window of the full far detector ensures that most muon tracks are contained in these images. Each pixel is filled with a value between 0 and 255 proportional to the cell's energy deposit with saturation at 278 MeV.

EventCVN classifies each event display image into one of (1) $\nu_\mu$ CC, (2) $\nu_e$ CC, (3) NC, and (4) cosmogenic background. The selections of $\nu_\mu$ and $\nu_e$ from data are independently generated with a series of data quality pre-cuts culminating in a cut on the EventCVN score of the respective class. Though not yet used in an oscillation analysis, NOvA also maintains ProngCVN \cite{prongcvn} for the purpose identifying individual particles, referred to as \textbf{prongs}. This network is trained on two images for each prong: the combined event pixel-map for contextual information and a prong pixel-map containing only the cell hits corresponding to an individual particle as determined by NOvA's prong reconstruction.

We introduce a hybrid convolution and transformer-based architecture known as TransformerCVN that makes use of NOvA's existing particle reconstruction to simultaneously process all particles within an event in a single network. This joint input scheme makes it possible to used contextual information to improve prong reconstruction and to describe the topological features of a particle's track that leads to its classification and to analyze the relative impact of each reconstructed particle on the overall event classification.

\section{Background}
\subsection{Attention and Set Classification}
Several problems in Physics may be reduced to assigning classification labels to a collection of unordered objects, or a set. In the case of NOvA event reconstruction, we are interested in classifying both the underlying event as well as reconstructing the label of each individual prong. Each event may contain multiple prongs, and these prongs have no inherent order to them. Therefore, the prong observations and targets define variable length sets and we may reduce the prong reconstruction task to classification over these sets.

There have recently been several developments in using attention-based methods to handle variable-length sets \cite{set_transformer, spanet, kamnet_network}. Attention provides a gating mechanism for modifying neural network activations by incorporating contextual information \cite{attention_baldi}. This technique has achieved state-of-the-art results in natural language processing problems such as translation \cite{attention_luong, attention_bahdanau, bert, gpt2}, where variable-length sequences are common. Attention has even found success in computer vision tasks by exploiting contextual similarities between image patches \cite{vit, swin}. Among these methods, transformers \cite{transformers} stand out as particularly promising for set assignment due to their fundamental permutation invariance \cite{set_transformer}. Transformers are especially effective at modeling variable-length sets because they can learn combinatorial relationships between set elements in polynomial time \cite{spanet}.

\subsection{Sparse Convolutions}
The scintillator pixel-maps produced by the NOvA detector present several unique challenges for machine learning. These pixel-maps are typically very sparse, with with events having, on average, 0.84\% of pixels containing non-zero hit values, leaving most of the observations void of data.

Convolution neural networks (CNNs) are ubiquitous element of deep learning methods for efficiently learning on image and other spatially-related collections of data \cite{deeplearning_goodfellow}. This effectiveness stems from the extreme weight sharing that comes with using a small \textit{kernel} matrix which is applied everywhere across space. However, the biggest successes of CNNs are on \textit{dense} images, such as photographs or artwork, due in part to the CNN's translation equivariance which allows them to learn spatial features such as edges or curves \cite{cnn_circuits}. 

CNN's begins to fail when dealing with sparse images since small kernels may not contain more than one non-zero element, and extremely large kernels would eliminate the benefits of weight sharing. To combat this, the spatial kernel concept has been expanded apply to sparsely distributed data, proving especially successful in 3D objective reconstruction \cite{submanifold, minkowski_nets}. These apply the convolution operations only in regions where data exists, saving on computation and preventing the dilution of sparse values even in a predominantly zero valued image. Additionally, this sparse convolution provides a method for generalizing convolutions to non-rectangular kernels, variable sized images, and additional modifications.

\subsection{Interpretable Deep Learning}
The black-box nature of deep neural network models stems from our inability to analytically describe the training and inference processes in all but the most simple neural networks \cite{deeplearning_goodfellow}. There has recently been a surge in methods for analyzing specific aspects of neural network architectures to extract human-understandable measurements from their internal structures. Saliency maps \cite{saliance} provide a method for analyzing the behaviour of CNNs by studying the model's output gradients with respect to the inputs. Saliency maps provide a very visual understanding of the network's behaviour near individual inputs. The convolution kernels of CNNs provide another method for extracting the \textit{features} learned by these models, and methods for analyzing these kernels provides further understanding of convolution operations \cite{gradcam, cnn_circuits}. Similarly, transformers may be analyzed by instead focusing on the attention matrices computed during self-attention \cite{transformers, transformer_circuits, bert}. These \textit{attention maps} measure the importance of different inputs, for example individual words in a language model, for determining the output of a transformer.

\section{Sparse Transformer CVN}
\begin{figure}[h]
    \centering
    \includegraphics[width=0.5\textwidth]{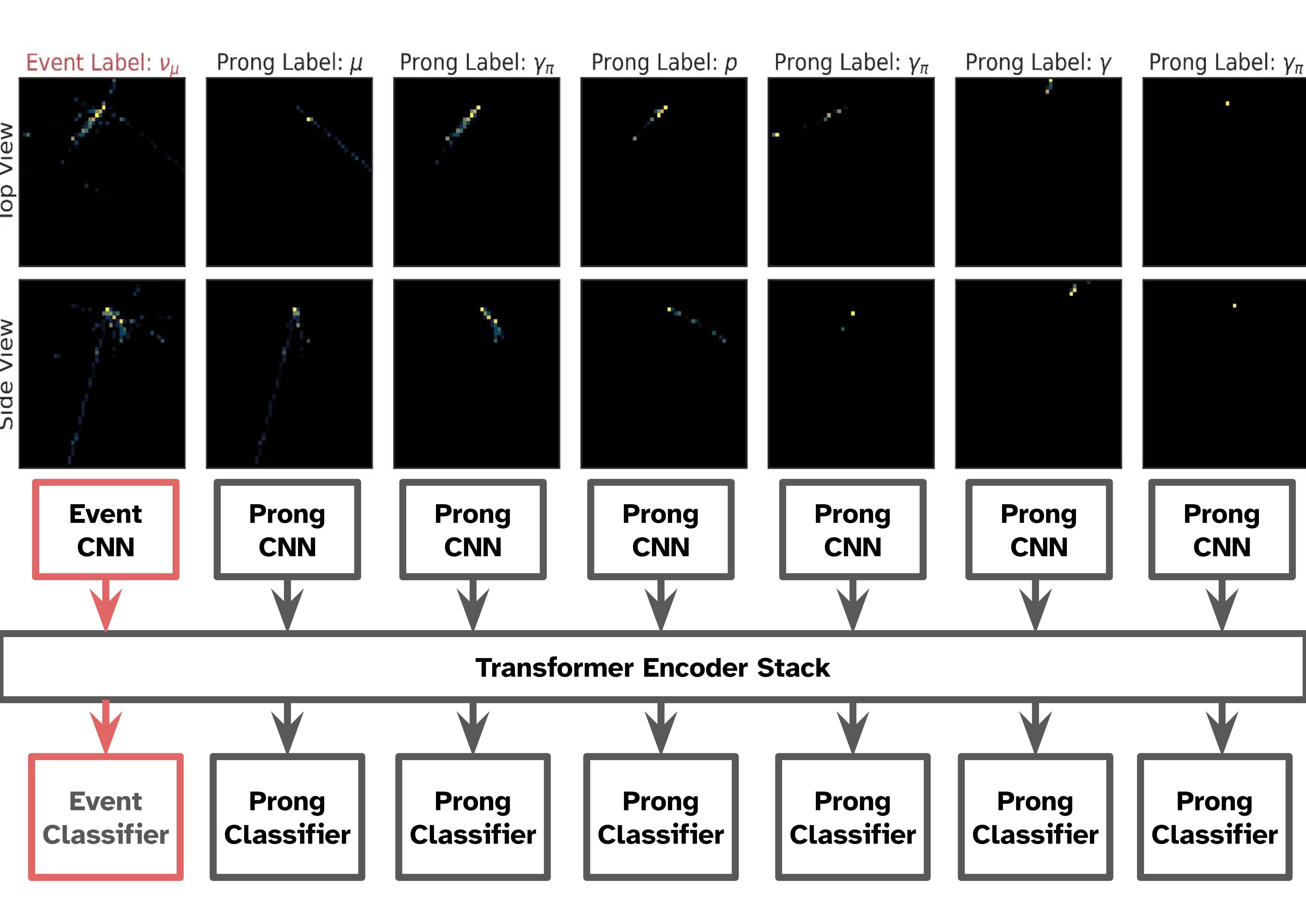}
    \caption{A complete diagram of Sparse Transformer CVN, including example pixel-maps from a $\nu_\mu$ event. The event pixel-map path is highlighted in red. The truth labels for each pixel-map are provided above each pixel-map.}
    \label{fig:network_diagram}
\end{figure}

We combine Sparse CNNs with Transformers in order to address both the pixel-map sparsity and the variable prong counts in each event. We introduce \textit{TransformerCVN} (Convolutional Visual Network), which embeds the sparse images into a dense latent space using Minkowski sparse CNNs \cite{minkowski_nets} before processing the the embeddings with transformer encoders \cite{transformers} to include contextual information. We provide as input both individual prong pixel-maps and a combined \textit{event} pixel-map since it may include hits which where not assigned to any particular prong. The individual prongs are associated their respective particle labels while the event pixel-map is associated with an overall neutrino interaction type. 


We use a densely connected residual CNN \cite{densenet} as the base architecture for the pixel-map embedding. This architecture introduces weighted skip connections between every pair of CNN layers, with the goal of ensuring that network activations do not decay due to input sparsity. We replace all of the traditional convolution and pooling operations in DenseNet with Minkowksi sparse convolution and pooling operations \cite{minkowski_nets}. We elect to only compute convolutions where the center pixel is non-zero, and we do not create additional non-zero values between sparse convolution operations, producing an architecture akin to a sparse sub-manifold CNNs \cite{submanifold}. This decision is motivated by the observation that our data typically consists of long traces, meaning the sparse convolution's receptive fields of will likely significantly overlap, even while only evaluating in non-zero regions. We embed the prong pixel-maps with a single CNN, sharing weights between prongs. A separately parameterized event CNN is used for event pixel-map to account for differences in its input distribution and classification objectives. Each pixel-map, originally two views at $100 \times 80$ pixels, is embedded into a single dense latent vector known as a \textit{embedded pixel-map}, with a size determined from a hyperparameter. The CNNs form the first stage of the network in Figure \ref{fig:network_diagram}.

We then processes these embedded pixel-maps with transformer encoder \cite{transformers} to allow contextual information to be shared between prongs. Both prong and event embedded pixel-map are \textit{encoded} using a single, shared transformer encoder stack, visible in Figure \ref{fig:network_diagram}. This transformer encoder follows the canonical formulation described in \cite{transformers}, with one major exception: we elect not the add position embeddings to our latent pixel-maps to avoid imposing an inherent order to the pixel-maps. We do, however, add a \textit{type} embedding to differentiate prong and event pixel-map by concatenating with one of two trainable vectors, one for prongs and one for events. This ensures the transformer is aware of the different objectives associated with these two inputs types while still allowing information to be shared between all components of an event.

\section{Joint Event-Prong Reconstruction Training}
\label{training}
After embedding and encoding the event and prong pixel-maps, the encoded vectors are fed through feed-forward networks to produce the final reconstruction predictions. We again use a single weight-sharing network for all prong reconstruction outputs and a separately parameterized network for event classification. The prong output networks produce softmax distributions over 9 possible prong targets, described in \ref{experiments} for every input prong. The event output network produces a softmax distribution over 10 possible event interaction types, and these outputs may be reduced to form a distribution over the four overall event types during inference.

We use a categorical log-likelihood loss to train both the prong reconstruction outputs and the event classifications, aggregating over all prong and event pixel-maps. Since every event has several classification targets and the distribution of targets is not perfectly balanced within the dataset, we add a focal loss term \cite{focal_loss} to our log-likelihood. This focal loss dynamically increases the weight of the events which the network is currently under-performing on. We find that adding this term improves reconstruction accuracy for secondary, non-leptonic prongs.

We use the AdamW \cite{adamw} optimizer to train TransformerCVN, and employ a cosine schedule with warm restarts for the learning rate \cite{cosine_annealing}. The AdamW optimizer along with warm-restarts has shown success on both vision \cite{yolov4, bag_of_tricks} and NLP \cite{nlp_ext} tasks with transformers. We replicate this setup in our experiments. We optimize the hyperparameters of our neural network using the Sherpa hyperparameter optimization framework \cite{hertel2020sherpa}. We use Bayesian Optimization with a Gaussian Process surrogate to guide the hyperparameter search process over $10,000$ short training trials. Final hyperparameters are presented in the Appendix.

\section{Experiments}
\label{experiments}

\paragraph{Dataset}
We train a Sparse Transformer CVN for both event-level and prong-level reconstruction on simulations of neutrino interactions in the NOvA far detector. One half of the simulated events consists of the unoscillated predominantly $\nu_\mu$ beam composition, and the other half consists of $\nu_e$ oscillated events. GENIE \cite{genie} and GEANT4 \cite{geant} were respectively used for the neutrino-nucleus interaction and detector simulation. The neutrino interactions are overlaid onto real cosmogenic background data. The dataset used in this analysis is the 5th production run of NOvA's Monte-Carlo simulation completed in 2020 \cite{simulation}. We then use the same pre-selection used to train EventCVN for NOvA's 2021 oscillation analysis \cite{oscillation}, consisting of NOvA's cosmic ray rejection veto and a cut on reconstructed transverse momentum fraction $p_T/p < 0.95$.

We split the dataset into training, validation, and testing splits in order to ensure low over-fitting. The training data consisted of $6,316,264$ events; the validation dataset, used for hyper-parameter optimization, had $332,434$ events; and a separately split testing dataset consisting of $177,084$ events was used to evaluate the model. The TransformerCVN and baseline models were trained and re-evaluated using the same set of splits. For storage efficiency, events were limited to 10 prongs with the lowest energy prongs removed for events exceeding 10 prongs. The final training events had a median prong count of 2, with a $90\%$ of events containing between 1 and 6 prongs.

\paragraph{Event Classification}
Event classification targets where assigned to be one of 10 possible labels $T_{event}$ =  \{$\nu_\mu$ CC QE, $\nu_\mu$ CC Res, $\nu_\mu$ CC DIS, $\nu_\mu$ CC Other, $\nu_e$ CC QE, $\nu_e$ CC Res, $\nu_e$ CC DIS, $\nu_e$ CC Other, $NC$, $CB$\} where QE is quasi-elastic scattering, Res is the resonant interaction, and DIS is deep inelastic scattering, $NC$ is neutral current events, and $CB$ is the cosmic ray background. However, we find that the network cannot effectively separate charged current events based on their interaction type. Due to this unreliable separation, and the fact that the EventCVN baseline did not include such a sub-classification, we collapse the event labels into these four basic classes for evaluation: $T_{eventCVN} = \{\nu_e, \nu_\mu, NC, CB\}$. We find the original dataset contained an overwhelmingly large number of cosmic background events. To prevent these events from skewing the networks predictions towards background, we down-sample $CB$ events to bring the distribution of event targets to be roughly uniform between the three signal classes, and limiting cosmics to only 10\% of the training dataset. Every event is assigned exactly one event label.

\paragraph{Prong Reconstruction}
Prong reconstruction targets were assigned from the possible set of $T_{prong} = \{ e, \mu, p, \gamma_n, \pi_{\pm}, \gamma_{\pi^0}, \gamma_{other}, OP, CB \}$. In an attempt to identify neutrons and neutral pions, which do not deposit energy in the scintillator, the photon class was split into $\{\gamma_n, \gamma_{\pi^0}, \gamma_{other}\}$ where $\gamma_{\pi^0}$ refers to a photon with a mother $\pi^0$. Non-cosmic prongs which did not fit into the more specific classes where given the other prong, $OP$, class. Prongs from data cosmic tracks do not have truth information and so were given the $CB$ class similarly to event reconstruction. Each discrete prong reconstructed from the events was assigned a single prong target. 

The ProngCVN baseline was not trained to separate the different mother particles for photons, or to identify the cosmic background prongs. Therefore, when comparing to ProngCVN, we use a simpler set of prong targets, $T_{ProngCVN} = \{ e, \mu, p, \gamma, \pi_{\pm} \}$. We derive this simpler labeling from TransformerCVN predictions by summing the different photon softmax outputs, dropping the unused labels, and re-normalizing the resulting classification distribution.

\paragraph{Performance}

\begin{figure}[h]
\centering
\begin{subfigure}{.32\textwidth}
  \centering
  \includegraphics[width=1.0\textwidth]{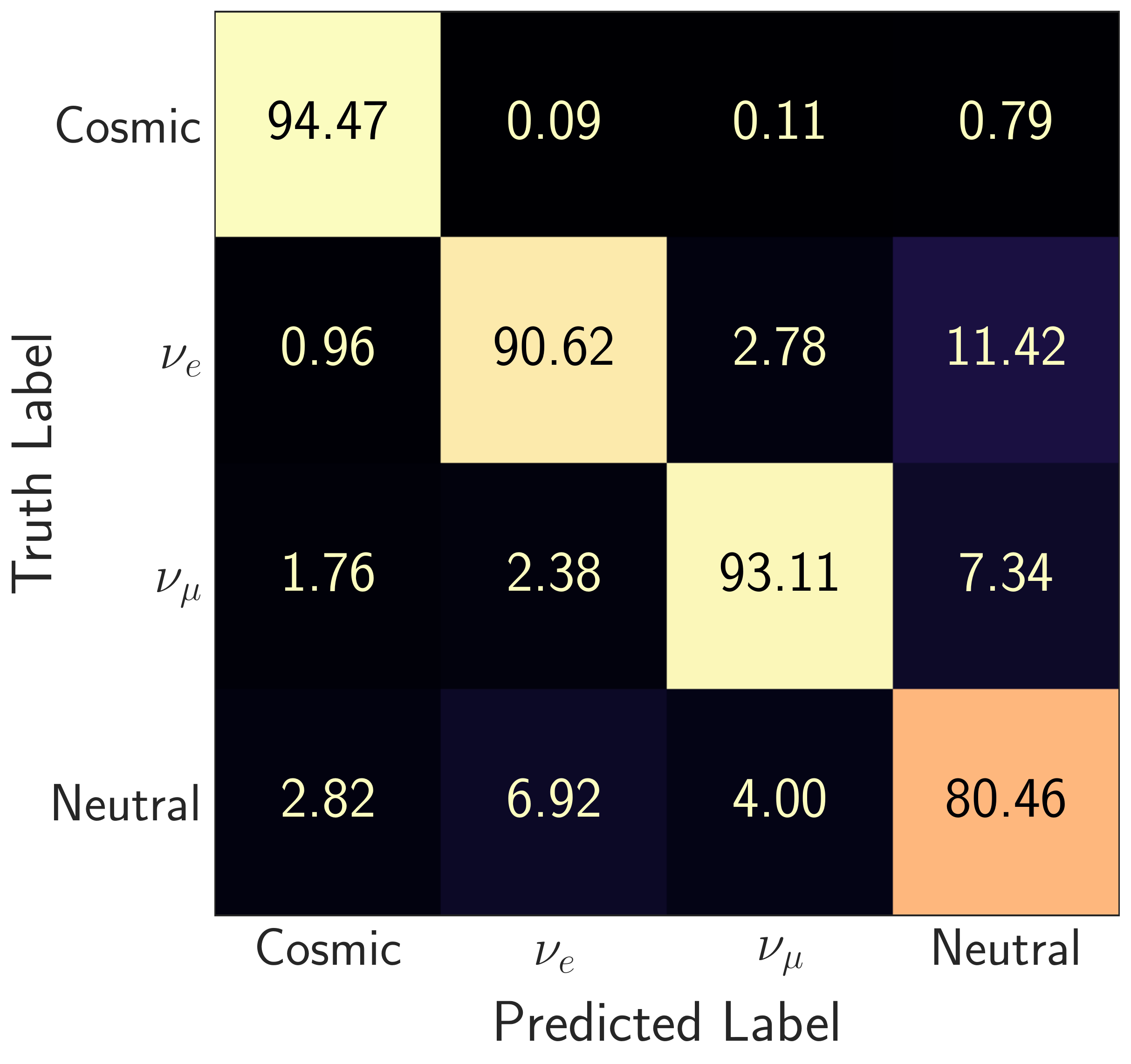}
  \caption{Event Reconstruction}
  \label{fig:event-confusion}
\end{subfigure}
\begin{subfigure}{.35\textwidth}
  \centering
  \includegraphics[width=1.0\textwidth]{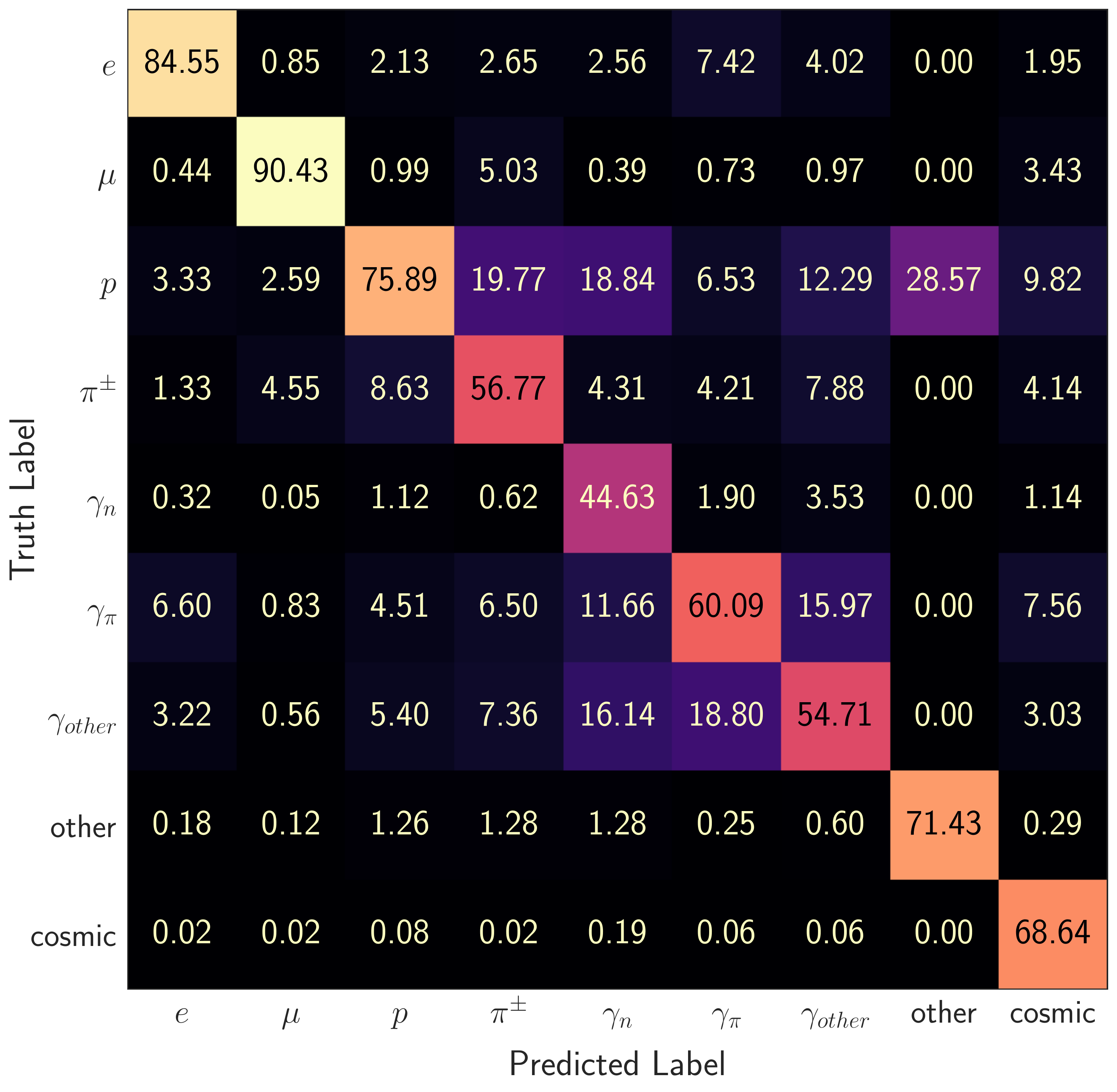}
  \caption{Prong Reconstruction}
  \label{fig:prong-confusion}
\end{subfigure}
\caption{TransformerCVN purity (prediction-normalized) confusion matrices.}
\label{fig:confusion}
\vspace{-16pt}
\end{figure}

\begin{table}[t]
    \hfill
    \parbox{1.0\linewidth}{
        \centering
        \begin{tabular}{c|cc}
            Metric & Transformer CVN & Event CVN \\
            \hline
            Accuracy & 0.894 & 0.897\\
            Precision & 0.894 & 0.908\\
            Recall & 0.894 & 0.897\\
            ROC AUC & 0.982 &  0.984\\
            \hline
        \end{tabular}
        \caption{Event reconstruction aggregated metrics}
        \label{tab:event_avg}
    }
    \hfill \\
    \vfill
    \parbox{1.0\linewidth}{
        \centering
        \begin{tabular}{c|cc}
            Metric & Transformer CVN & Prong CVN \\
            \hline
            Accuracy & 0.783 & 0.726\\
            Precision & 0.783 & 0.760\\
            Recall & 0.783 & 0.726\\
            ROC AUC & 0.951 &  0.932\\
            \hline
        \end{tabular}
        \caption{Prong reconstruction aggregated metrics.}
        \label{tab:prong_avg}
    }
    \hfill
    \vspace{-16pt}
\end{table}

We evaluate the baseline reconstruction performance on on both event and prong reconstruction for TransformerCVN. We present average values across the testing dataset for various metrics in Tables \ref{tab:event_avg} and \ref{tab:prong_avg}. We average the metrics in a one-versus-one approach, re-weighting the classes for a balanced metric. 

Additionally, we compute a confusion matrix with respect to the truth label for both event and prong reconstruction tasks. We present confusion matrices for both event and prong reconstruction in Figure \ref{fig:confusion}. We also normalize these truth tables over both true targets (Efficiency) and predicted targets (Purity), presenting all confusion matrices for both TransformerCVN and baselines in the Appendix. For the event reconstruction, we notice that most of the classes are well identified, with the major confusion coming from separating $\nu_e$ CC and NC events. In prong reconstruction, we notice that protons where the most miss-classified auxiliary prong type, sharing a similar profile to the different types of photons. We additionally notice that the network has trouble separating the photons by source, but can typically still determine a general photon class with high purity.

\paragraph{Baselines}
We compare the Sparse Transformer CVN (Labeled \textit{TransformerCVN} in plots) to two baseline models currently used for NOvA reconstruction. We compare the event current prediction against the original NOvA EventCVN \cite{cvn}. We compare the prong reconstruction performance against ProngCVN \cite{prongcvn} which uses only the current prong and an event pixel-map. ROC Curves for comparison are provided in the Appendix (Figures \ref{fig:event_roc_curves} and \ref{fig:prong_roc_curves}). 

We notice that event performance remains nearly identical to the event CVN model, with one small benefit of TransformerCVN appearing to produce a more balanced prediction with identical precision and recall. This likely means that both of models are near-peak performance for the given dataset. However, this also means that we do not experience any downsides on this baseline reconstruction result by adding additional prong reconstruction objectives. 

TransformerCVN shows great improvement in prong reconstruction. The additional context provided by all prongs and the transformer's attention mechanism improves prong AUC across all of the major particle types when compared to the ProngCVN. We achieve an 0.02 increase in a one-versus-one pairwise aggregated AUC when compared to ProngCVN and an increase in nearly 5\% reconstruction accuracy. This improvement is most pronounced on the critical lepton prong reconstruction, where we experience an improvement in $\mu$ prong AUC from 0.864 to 0.975 (Figure \ref{fig:prong_roc_curves}).

\begin{figure}[h]
\centering
\begin{subfigure}{.48\textwidth}
  \centering
  \includegraphics[width=1.0\linewidth]{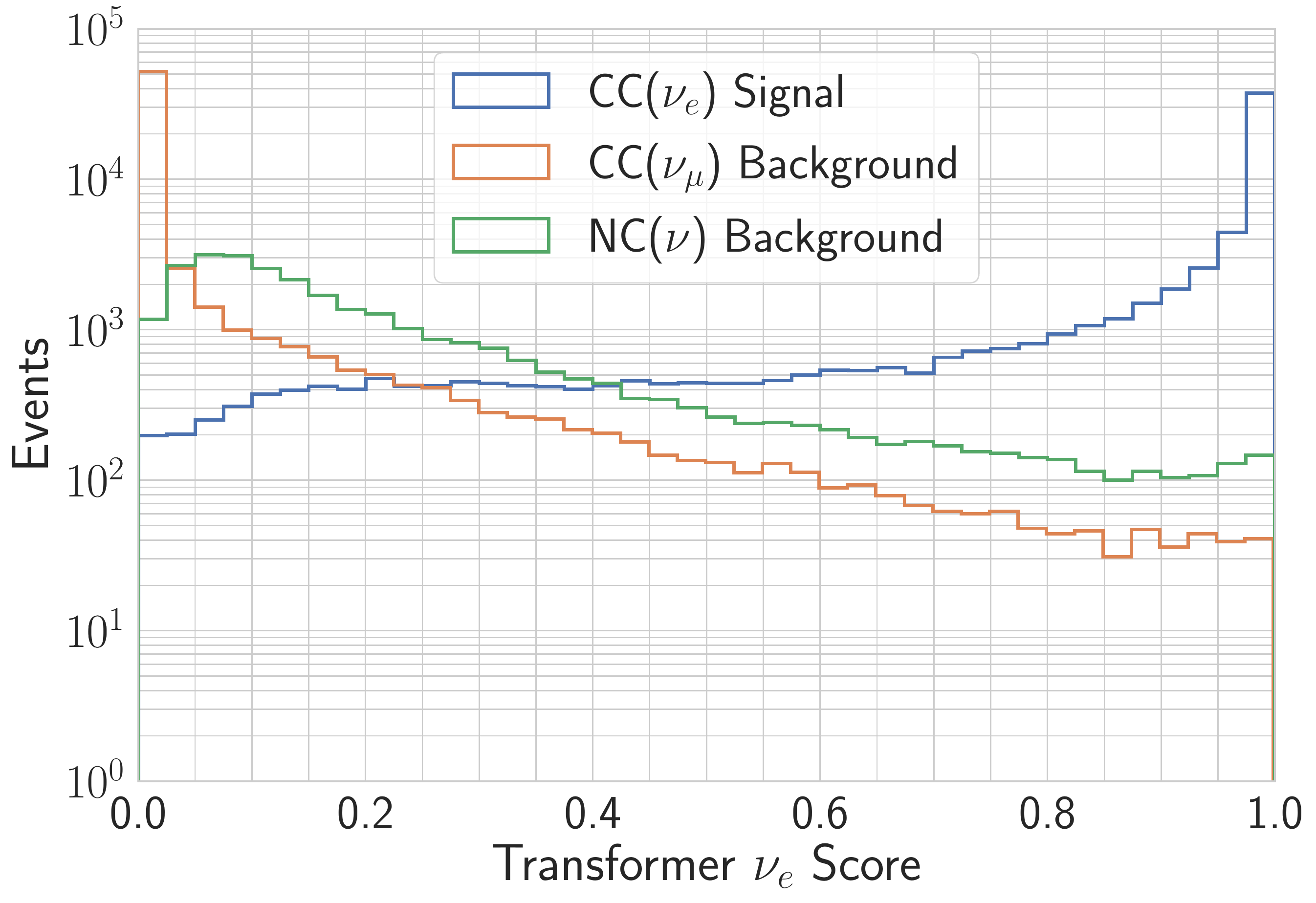}
  \caption{Transformer $\nu_e$ Event Softmax Scores}
\end{subfigure}%
\hfill
\begin{subfigure}{.48\textwidth}
  \centering
  \includegraphics[width=1.0\linewidth]{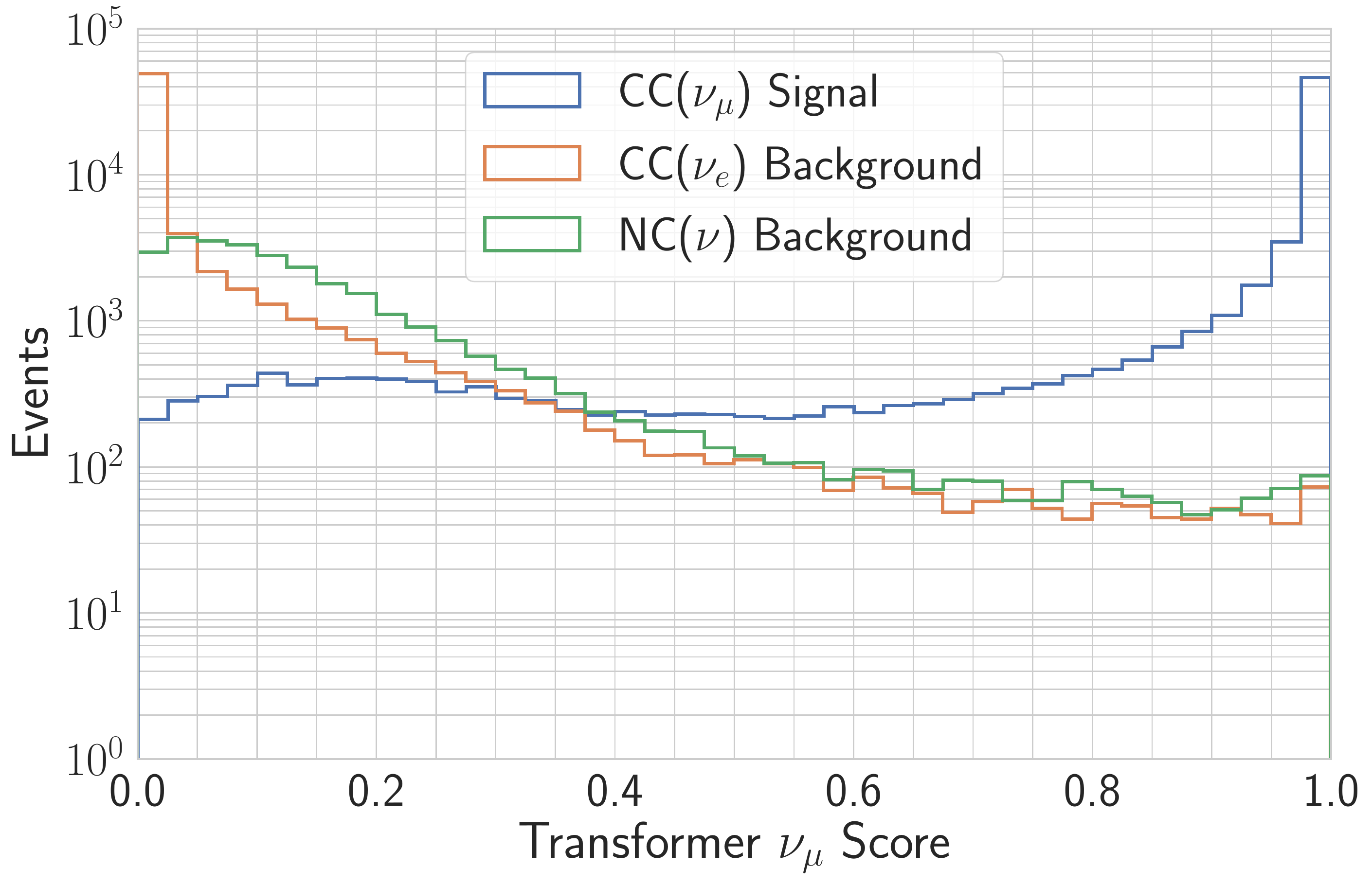}
  \caption{Transformer $\nu_\mu$ Event Softmax Scores}
\end{subfigure}
\caption{Event signal-background rejection curves for different Event Types. Calculated as the TransformerCVN's likelihood of classifying a particular signal event as one of the background classes.}
\label{fig:event-signal-background}
\end{figure}

\begin{figure}[h]
\centering
\begin{subfigure}{.48\textwidth}
  \centering
  \includegraphics[width=1.0\linewidth]{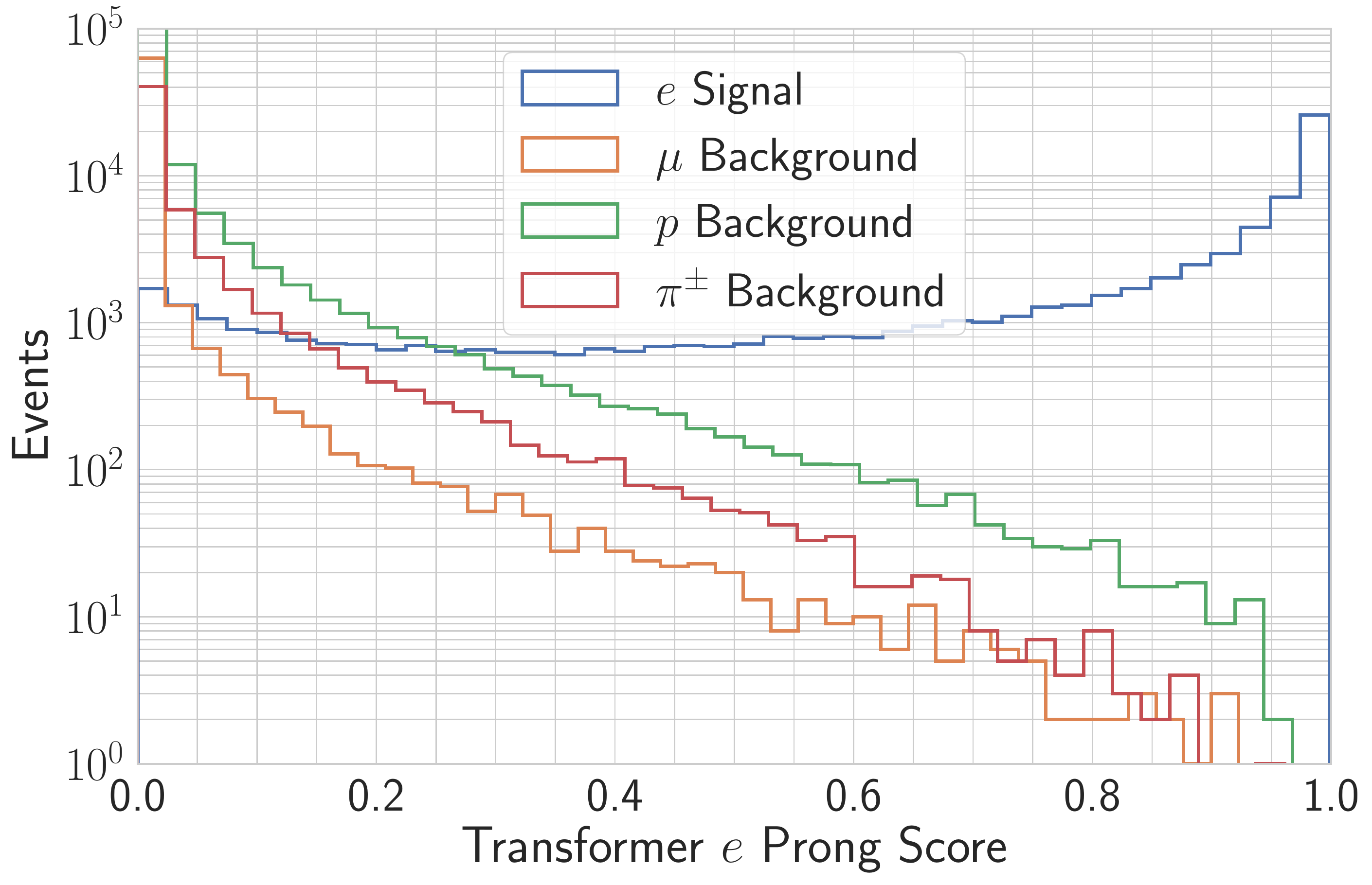}
  \caption{TransformerCVN $e$ Prong Softmax Scores}
\end{subfigure}%
\hfill
\begin{subfigure}{.48\textwidth}
  \centering
  \includegraphics[width=1.0\linewidth]{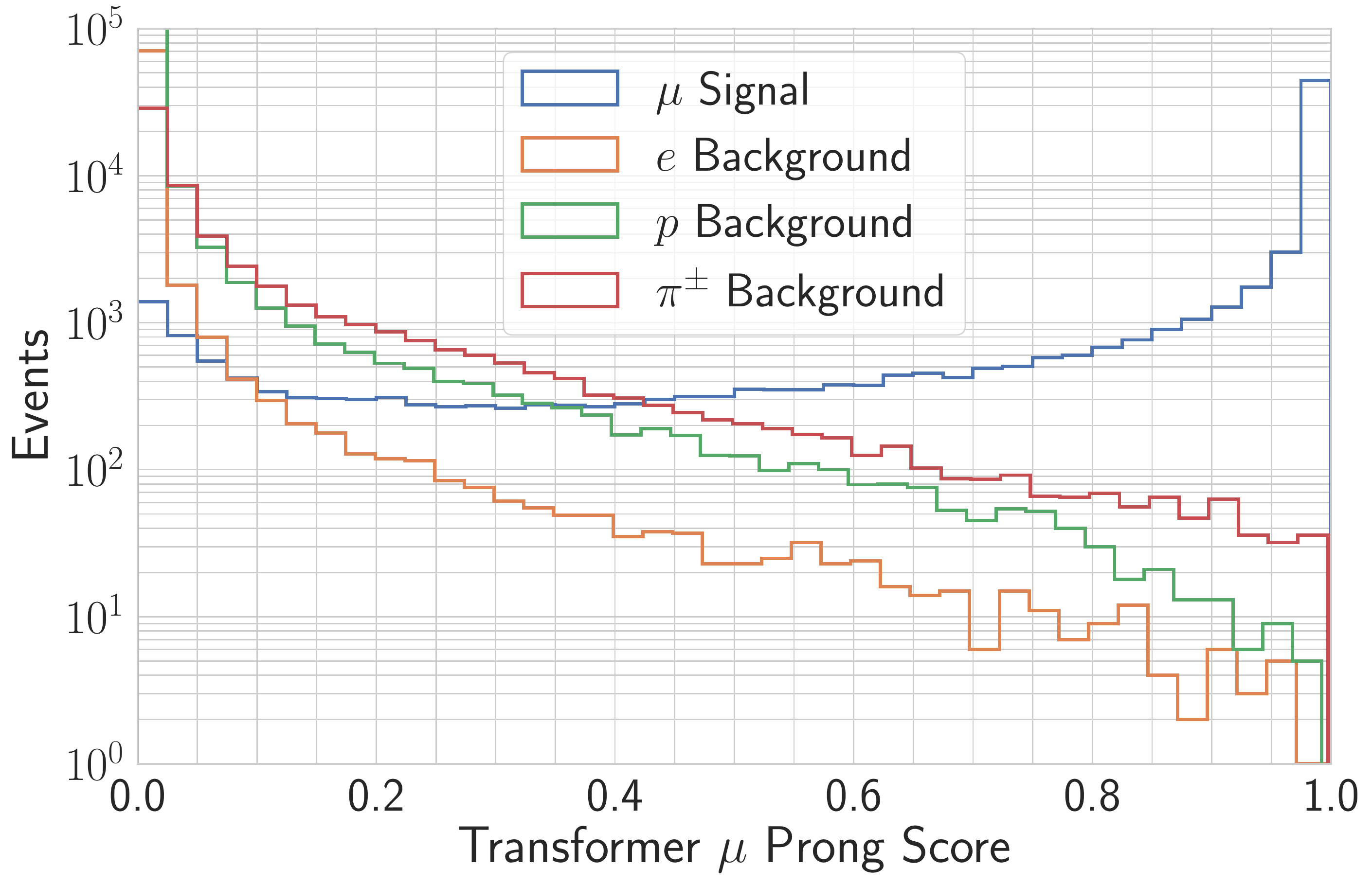}
  \caption{TransformerCVN $\mu$ Prong Softmax Scores}
\end{subfigure}
\caption{Prong signal-background rejection curves for lepton prongs. Equivalent computation as the event signal-background curves, but performed with the four most common types of prongs.}
\label{fig:prong-signal-background}
\end{figure}

\paragraph{Signal Background Rejection}
We can also use the softmax distribution outputs from both the event and prong network outputs as a method for cutting non-signal events from the NOvA data. To examine the effectiveness of this cut, we plot a histogram of the network's softmax probability of assigning a given classification for every event or prong, grouped by their ground truth values. These plots are presented in Figures \ref{fig:event-signal-background} and \ref{fig:prong-signal-background}. We notice that all of the major events and prongs achieve an order of magnitude signal-background cut after a classification probability of $0.8$ while still keeping a majority of the signal data with a cut up to $0.9$.

\section{Interpretability}
One often cited disadvantage of deep neural networks is the degree to which their complex architectures are black boxes, offering no explanation behind their predictions. In spite of the the enormous number of learned parameters, the simple and unified structure of the TransformerCVN offers several opportunities to "open up" the black box.
By tracking the attention mechanism throughout the network, we build an interpretable understanding of the relationships between different particles and event types. We further reverse engineering the spatial structures learned by the CNNs to construct spatial profiles of different particles, providing insight into methods of separating similar particles. Through these interpretability studies, we find evidence suggesting the network learns several known principles from the standard model.

\paragraph{Attention Maps}
\begin{figure}[t]
\centering
\begin{subfigure}{.48\textwidth}
  \centering
  \includegraphics[width=1.0\linewidth]{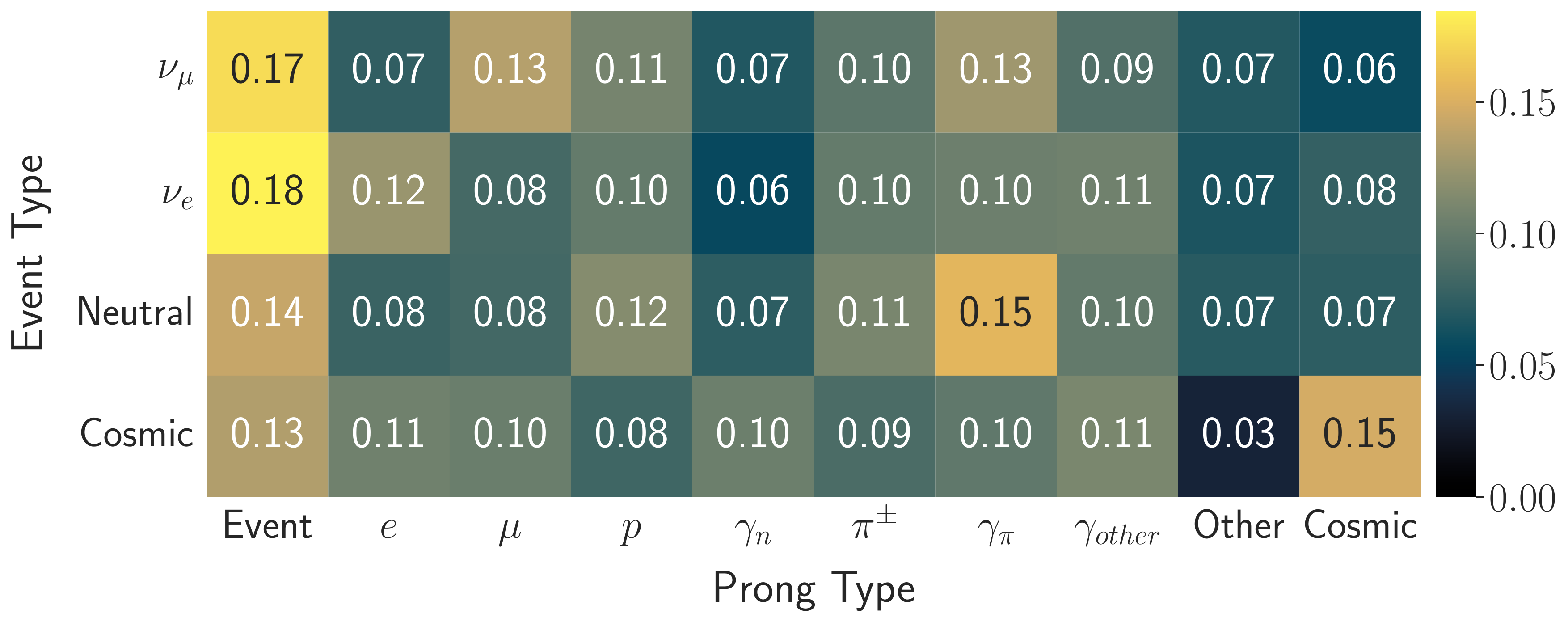}
  \caption{Event attention scores.}
  \label{fig:attention-event}
  \vfill
\end{subfigure}%
\hfill 
\\
\vfill
\begin{subfigure}{.48\textwidth}
  \centering
  \includegraphics[width=1.0\linewidth]{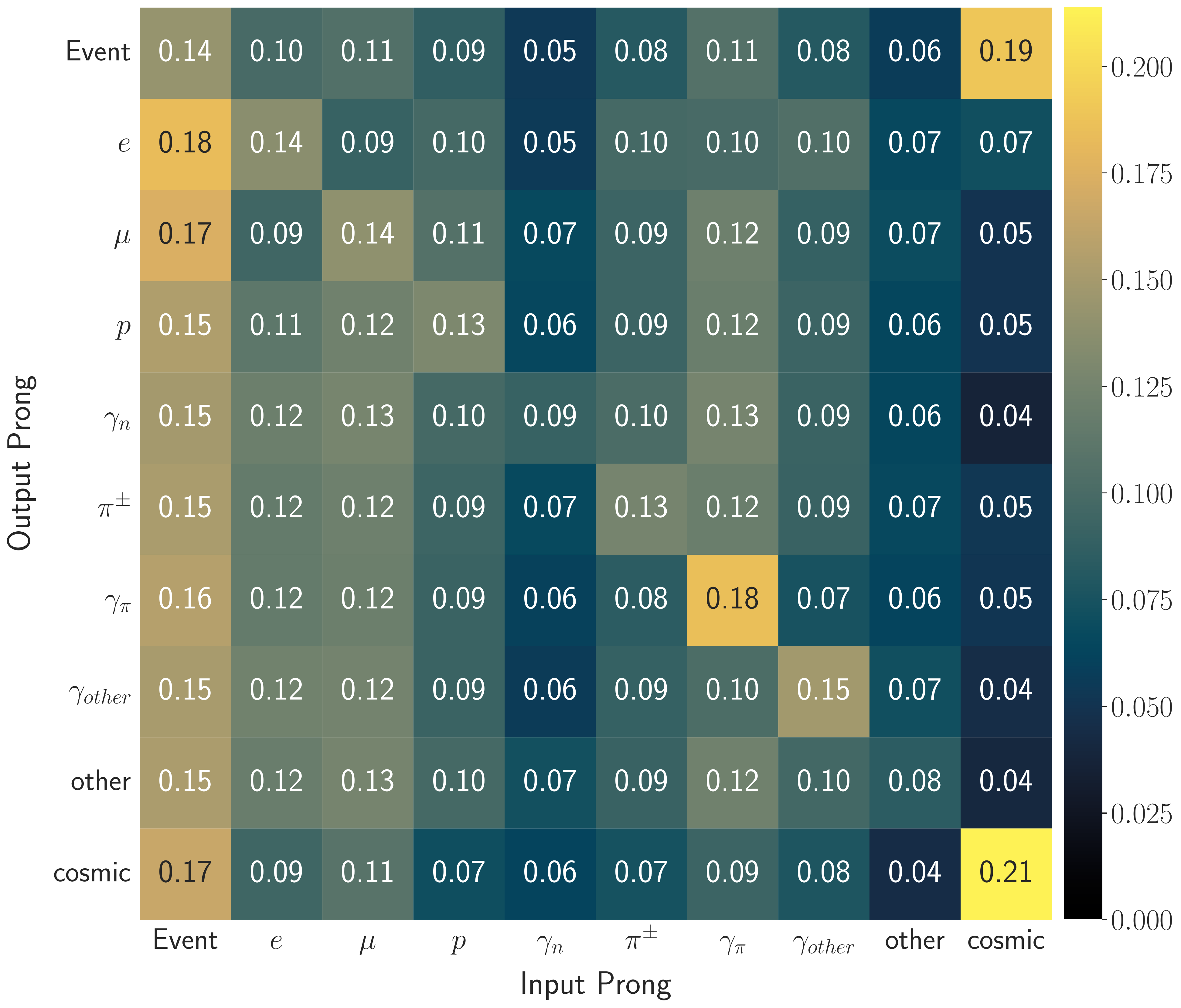}
  \caption{Prong pairwise attention scores.}
  \label{fig:attention-prong}
  \vfill
\end{subfigure}
\caption{Aggregated attention matrices measuring the impact of different prong types to various predictions.}
\label{fig:attention}
\vspace{-6pt}
\end{figure}

One major advantage of using networks based on attention such as transformers is that we may easily visualize the attention weights attributed to each input. This provides a numerical importance of each input to each output, providing an indication of how much the particular input contributed to the given output. We may compute the total attention across the entire encoder stack by simply taking the product of of individual attention maps.

Let $E \in \mathbb{R}^{(N + 1) \times D}$ be a set of embedded pixel-maps where $N$ is the number of prongs in the event and $D$ is the latent dimensionality. Every transformer layer, $T_i$, in a $K$-layer transformer encoder produces a pair-wise importance score for all of the input pixel-maps: $A_i \in\mathbb{R}^{(N + 1) \times (N + 1)}$. We may compute a total attention score for a single event by taking the product of these importance matrices across the entire transformer encoder.
$$ A = A_K A_{K - 1} \dots A_2 A_1 \mathbb{I} $$
This produces a single $(N + 1) \times (N + 1)$ matrix who's rows all sum to one. We note that high attention values does not indicate correlation with predicting the given output, but rather the \textit{importance} for separating different classifications.

We require a special procure to aggregate these attention maps between events due to the different prong counts in each event. We use each prong's truth labels as the \textit{type} of said prong, and we include the additional event pixel-map as an \textit{event} type pseudo-prong for the purposes of this analysis. Since events have variable prong counts, we use the logit-attention values, $\log \left ( A \right )$, instead of the softmax attention scores. We aggregate these attention maps by first finding the average logit-attention score for every prong type, summing the attention score for all prongs of the same type in each event, before finding the average of the total logit-attention for that type. This presents the importance of each prong type to the reconstruction of each prong type while accounting for the typically higher number of secondary prongs within an event compared to leptonic prongs. We present these aggregated pairwise attention scores in Figure \ref{fig:attention-prong}. We see a generally diagonal patter, with the exception of event pixel-map inputs which have larger attention scores than the individual prongs.


We also take a deep dive into the attention scores for individual event outputs. We perform a similar aggregation technique over just the event prong's row in the attention matrix, grouping the attention scores based on the true event label. This allows us to examine the importance of different types of prongs for providing the context necessary to make an event-level prediction. We present this analysis in Figure \ref{fig:attention-event}. As expected, the presence of particles largely unique to certain interaction types have large impacts namely elections for $\nu_e$ CC, muons for $\nu_{\mu}$ CC, and neutral pions for NC.

This novel form of aggregation allows us to build a more general understanding of the network's predictions, avoiding the pitfalls of typically single example-based experiments of interpretability. We also perform alternate aggregations in a per-prong fashion, ignoring the fact that events may have more than one prong of the same type in each event. This generally produces less informative attention maps, and we present this analysis in the Appendix.

\paragraph{CNN Saliency Maps}

\begin{figure}
\centering
\begin{subfigure}{.48\textwidth}
  \centering
  \includegraphics[width=1.0\linewidth]{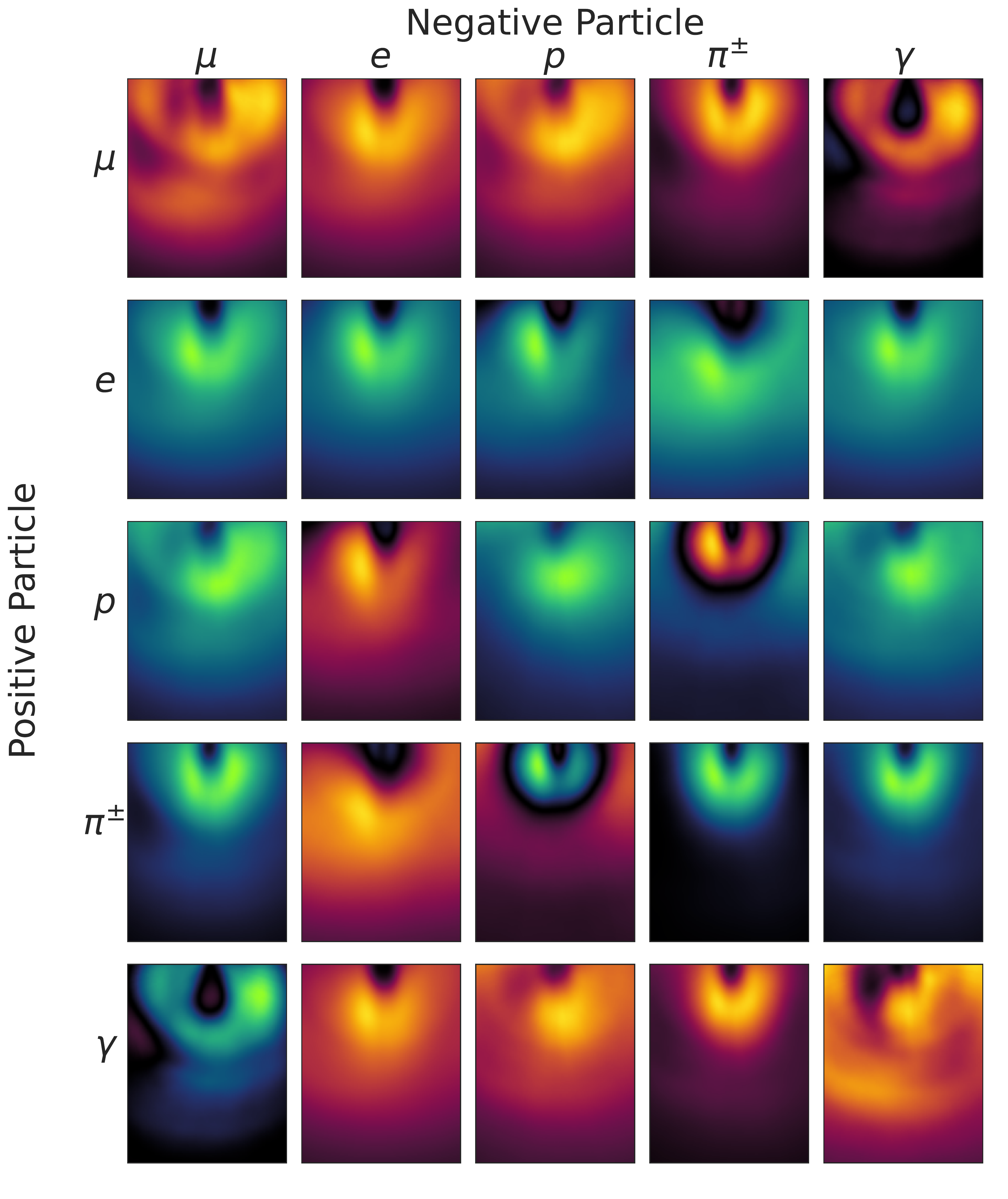}
  \caption{All prongs examined in every saliency map.}
  \label{fig:saliency_all}
  \vfill
\end{subfigure}%
\hfill
\begin{subfigure}{.48\textwidth}
  \centering
  \includegraphics[width=1.0\linewidth]{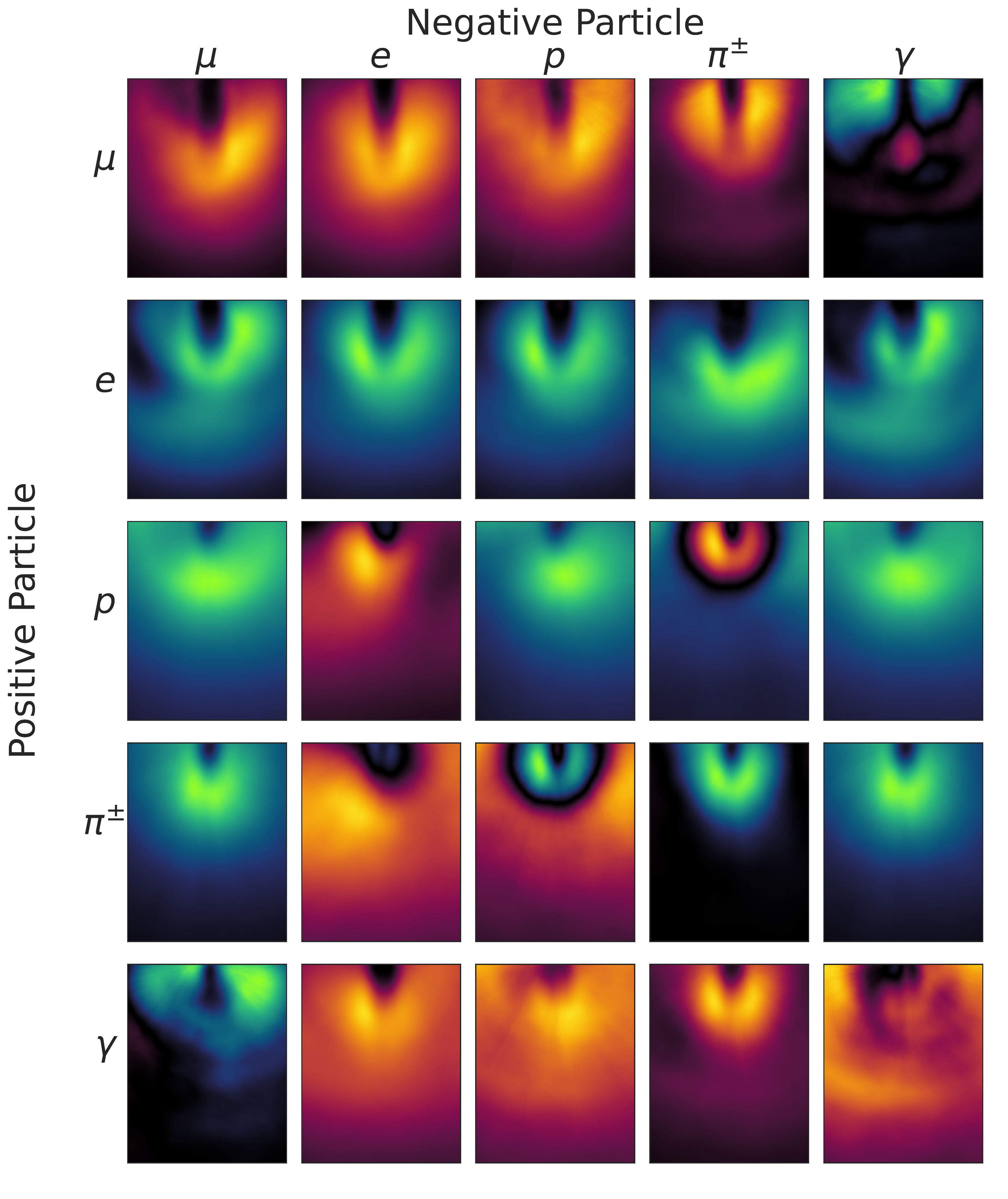}
  \caption{Each row only contains prongs who's truth label matches the positive particle for the given row.}
  \label{fig:saliency_target}
  \vfill
\end{subfigure}
\caption{Grid of aggregated saliency maps and difference maps for every pair of prong types. Red indicates a positive correlation with the Positive Particle's reconstruction probability, while blue represents a negative correlation.}
\label{fig:saliency}
\end{figure}

One method for interpreting the learned behaviour of convolution networks is to use saliency maps \cite{saliance}. Saliency refers to the derivative of a network output with respect to the input pixel, $\frac{\partial O}{\partial I}$. This produces pixel-maps, with identical shape to the input, representing how strongly the output probability will change with respect to an infinitesimal change in intensity for every input pixel. We produce these saliency maps for every prong's output and event output with respect to all input pixel-maps, both prong and event. Examples of complete saliency maps for individual events are presented in the supplementary materials. 

Saliency maps are typically very noisy for individual events, especially due to our input sparsity. Therefore, we present a method for aggregating saliency maps across multiple events to extract an average, interpretable result for understanding the underlying physics. We perform this aggregation in several steps.
\begin{enumerate}
    \item Compute the saliency of every output head for several blurred, noisy variations of each prong. We add a small uniform random noise to pixel-maps to robustly estimate the gradient near a given input, followed by a Gaussian blur with a standard deviation of 1 pixel to smooth out the discrete nature of our pixel-maps. We average the saliency across these noisy blurred inputs to provide a smooth gradient estimates. This produces 9 saliency maps for every prong, one for each prong label.
    \item Translate and rotate each saliency map to align each prong's vertex (the initial location of the hit) to the top center of each map and the prong's track (the decay tail) along the vertical axis. We use vertex and direction information from the simulator's particle reconstruction, included in the NOvA MC data release.
    \item Enforce a similar distribution for every prong type by limiting events to only those where the track length is less than 50 pixels (488 cm) and the reconstructed particle energy is less than 4 GeV. We do this to focus on the differentiation of similar-looking prongs, providing hints to subtle differences between different prong types instead of obvious overarching differences.
    \item Average the resulting smoothed and rotated saliency maps for every type of prong to compute the gradient with respect to deflection way from the vertex for each prong type.
\end{enumerate}

We present several grids displaying these saliency maps for five major prong particle labels in Figure \ref{fig:saliency_all}. The diagonal pixel-maps present saliency maps for the particle type associated with that diagonal element. The off-diagonal maps presents pair-wise difference saliency maps: computed as simply $\left ( \textit{Positive Particle} - \textit{Negative Particle} \right )$. These maps indicate what regions correlate the most with the network predicting the prong is more likely to be the Positive Particle label rather than the Negative Particle. Figure \ref{fig:saliency_target} displays a similar visualization, but limiting prongs present in each row to only those who's truth label match the \textit{Positive Particle}. This produces an asymmetric grid since every row now contains only those prongs that match the row's label.

One would expect that muons are the easiest to visually identify from the other particle classes considered here due to their tendency to leave tracks rather than showers. It is clear from the muon rows of both figure \ref{fig:saliency_all} and \ref{fig:saliency_target} that hits near the vertex make the network more likely to classify the prong as a muon track as opposed the shower of a showering particle. $\mu/\gamma$ separation in figure \ref{fig:saliency_target} presents an interesting example where hits at large angles from the prong direction as opposed to hits along the direction vector far from the vertex make $\gamma$ classification more likely. $\gamma$ separation from $e$ usually relies on a predicted gap between the vertex and the start of the photon's shower. Hits far the vertex therefore contribute more to $\gamma$ classification. However, the $e/\gamma$ separation plots appear more isotropic than $\mu/\gamma$ or $\mu/e$ plots, likely because both types of particles are expected to shower.

\paragraph{Integrated Saliency Maps}

\begin{figure}
\centering
\includegraphics[width=1.0\linewidth]{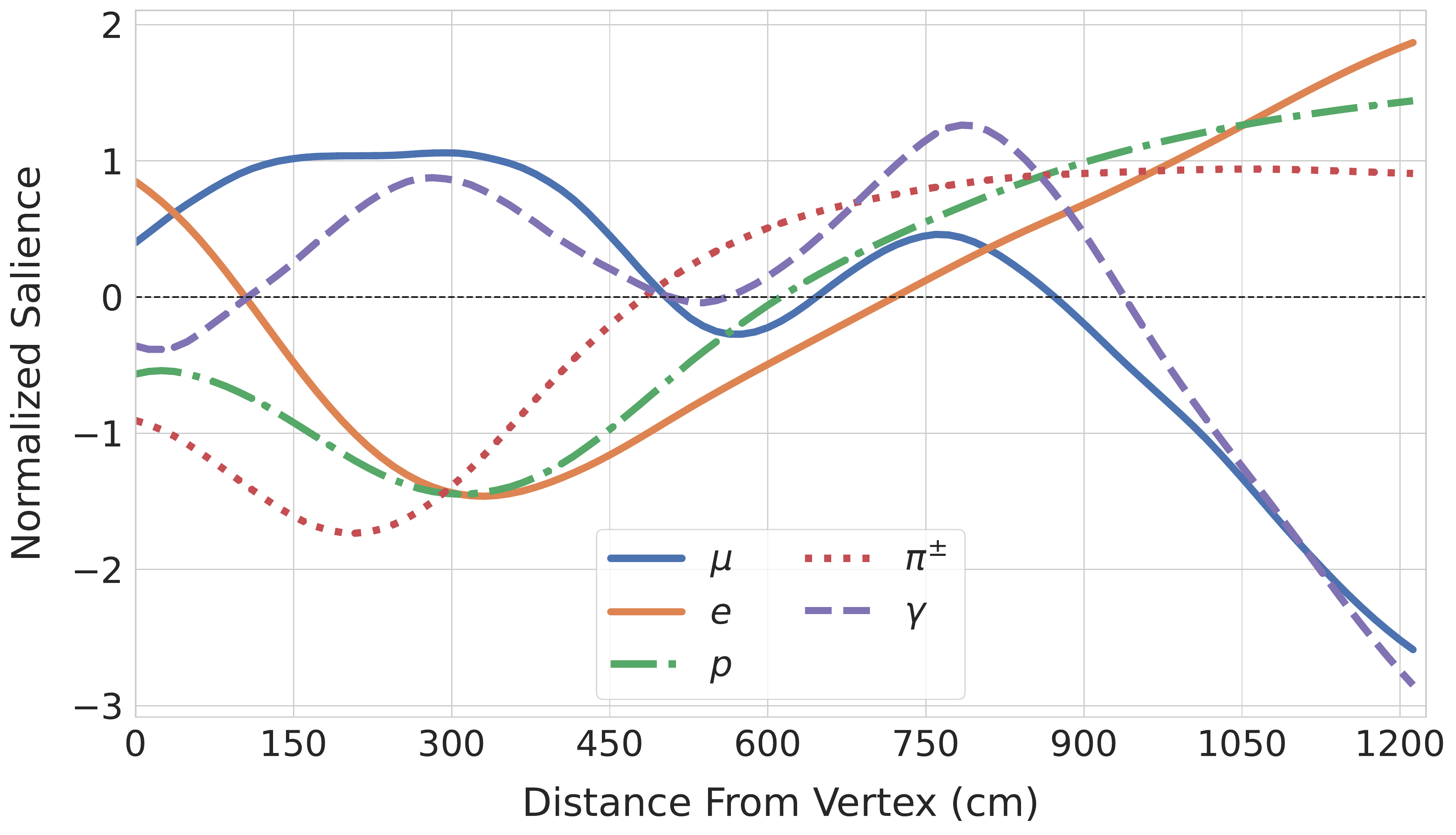}
\caption{Integrated salience map demonstrating the trace profile of different prong types.}
\label{fig:integrated-saliency-example}
\vspace{-12pt}
\end{figure}

We may also compare the track profile between the different classes by integrating the saliency maps presented above across the width of the detector to produce one-dimensional saliency with respect to vertex distance along the track (Figure \ref{fig:integrated-saliency-example}). This provides an average "pattern" we expect every prong type to form and indicates the most important regions for each prong class. We again notice that the muon track extends further than the other prong types, remaining flat for the middle third of the track. We also notice the delayed hit expected from $\gamma$ tracks when compared to $e$.


\section{Conclusion}
We present a novel neural network architecture for event reconstruction at NOvA. By combining the spatial correlation learning of sparse convolution networks with the contextual learning of transformers, we present a method for simultaneously reconstructing both individual prong labels as well as an overall event classification. This combined approach improves reconstruction accuracy over baseline methods while also providing many novel methods for interpreting the network's reasoning behind individual reconstructions. The black-box nature of neural networks is often an uneasy aspect of deep learning models, we a method for "opening the black box" provides a method for increasing trust in the network's predictions and guides our understanding of the underlying physics. Interpretable networks are critical for not just improving current physics experiments, but for guiding our understanding in designing new experiments.


\bibliographystyle{unsrt}
\bibliography{bib}

\section{Appendix}

\subsection{Hyperparameters}
We present a full list of hyperaprameters used to define the network. The CNN follows the DenseNet architecture \cite{densenet} with a modified number of blocks and embedding dimensions. The prong transformer follows the canonical transformer encoder \cite{transformers} architecture. We used a focal classification loss \cite{focal_loss} with a chosen focal $\gamma$ parameter. AdamW \cite{adamw} and cosine annealing with warm restarts \cite{cosine_annealing} with canonical parameters are used for training the network. Training was performed on 4 NVidia 3090 GPUs, splitting a batch size 2048 events between the GPUs.

\begin{table}[h]
    \centering
    \begin{tabular}{l|r}
        \textbf{Parameter} & \textbf{Value} \\
        \hline
        CNN Embedding Dimensions & $512$ \\
        CNN DenseNet Blocks & $5$ \\

        Transformer Dimensions & $256$ \\
        Transformer Encoders & $6$ \\
        
        Type Embedding Dimensions & $32$ \\
        
        Focal Loss $\gamma$ & 1.0 \\
        AdamW Learning Rate & $1 \times 10^{-5}$ \\
        AdamW Weight Decay & $2 \times 10^{-5}$ \\
        Cosine Annealing Epochs & $1024$ \\
        Cosine Annealing Cycles & $16$ \\

    \end{tabular}
    \caption{Caption}
    \label{tab:my_label}
\end{table}

\subsection{Detailed Plots}


\begin{figure}[h]
\centering
\begin{subfigure}{0.45\textwidth}
  \centering
  \includegraphics[width=1.0\linewidth]{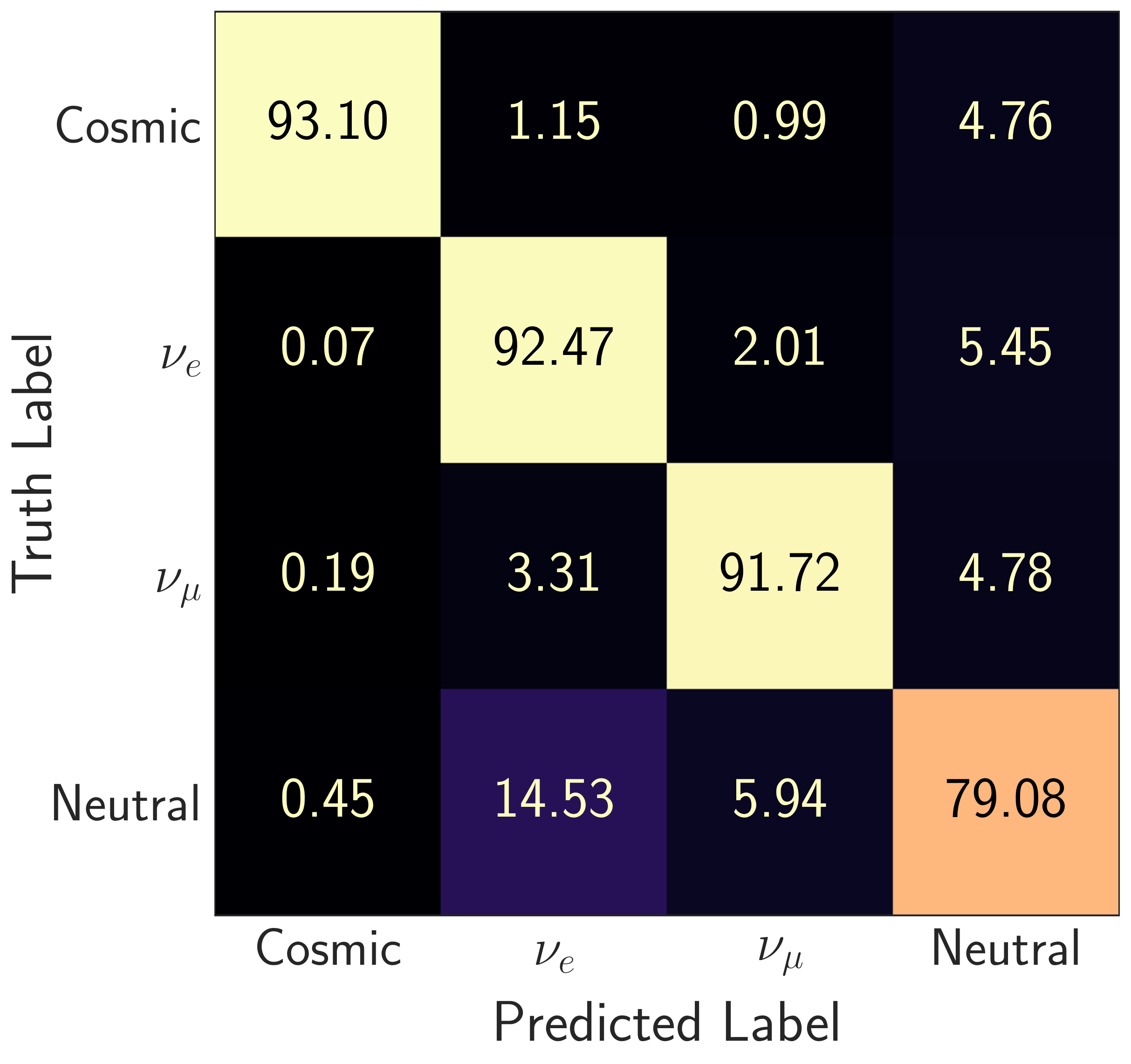}
  \caption{\textbf{Efficiency} matrix, normalized along truth labels.}
  \label{fig:event-confusion-efficiency-appendix}
\end{subfigure}%
\hfill
\begin{subfigure}{.45\textwidth}
  \centering
  \includegraphics[width=1.0\linewidth]{Figures/Confusion/nova_event_targets_confusion_pred.pdf}
  \caption{\textbf{Purity} matrix, normalized along predictions.}
  \label{fig:event-confusion-purity-appendix}
\end{subfigure}
\hfill
\caption{TransformerCVN 4 Class Event Confusion Matrices.}
\label{fig:event-confusion-appendix}
\end{figure}

\begin{figure}[h]
\centering
\begin{subfigure}{.45\textwidth}
  \centering
  \includegraphics[width=1.0\linewidth]{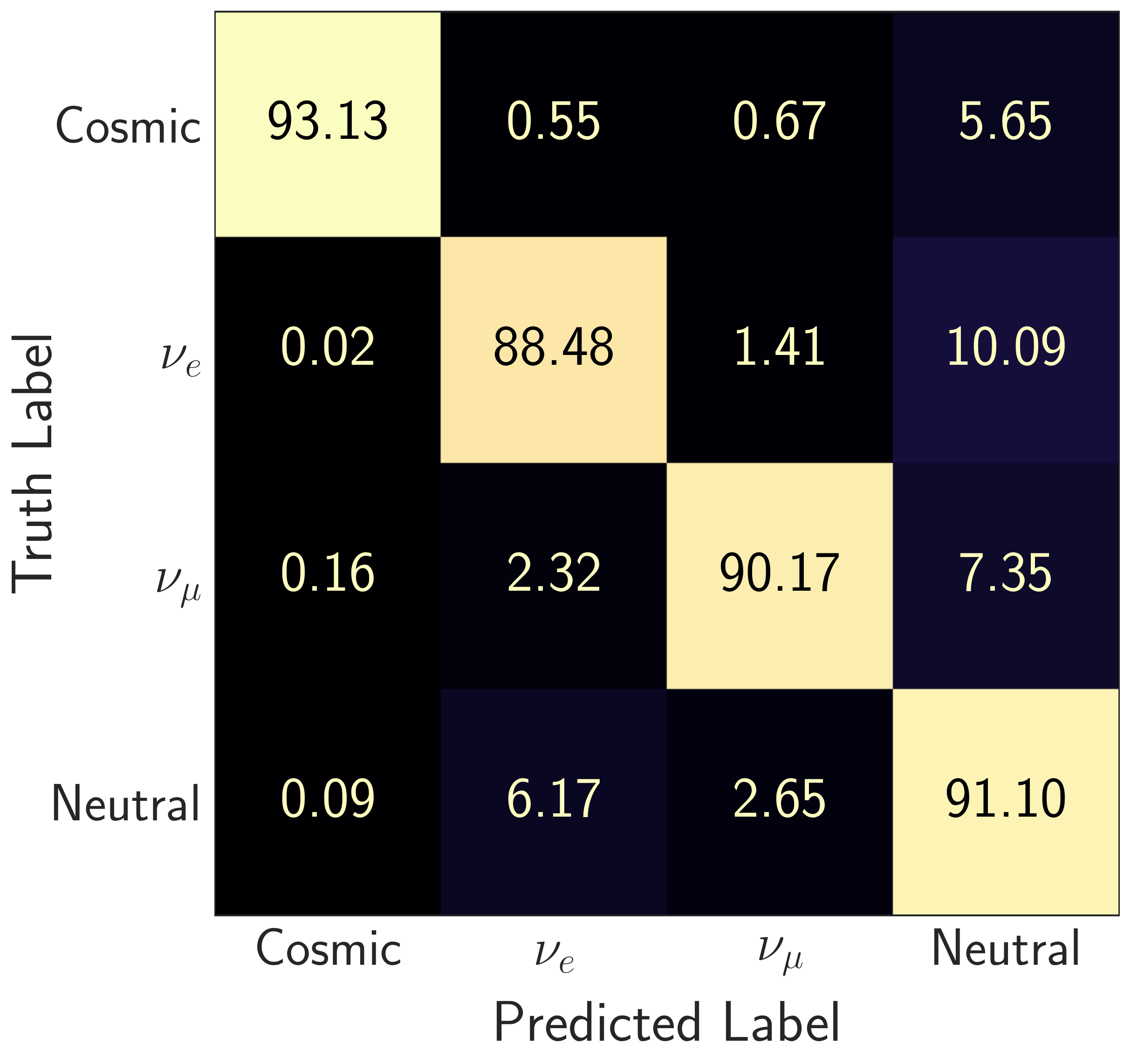}
  \caption{\textbf{Efficiency} matrix, normalized along truth labels.}
  \label{fig:event-confusion-efficiency-baseline-appendix}
\end{subfigure}%
\hfill
\begin{subfigure}{.45\textwidth}
  \centering
  \includegraphics[width=1.0\linewidth]{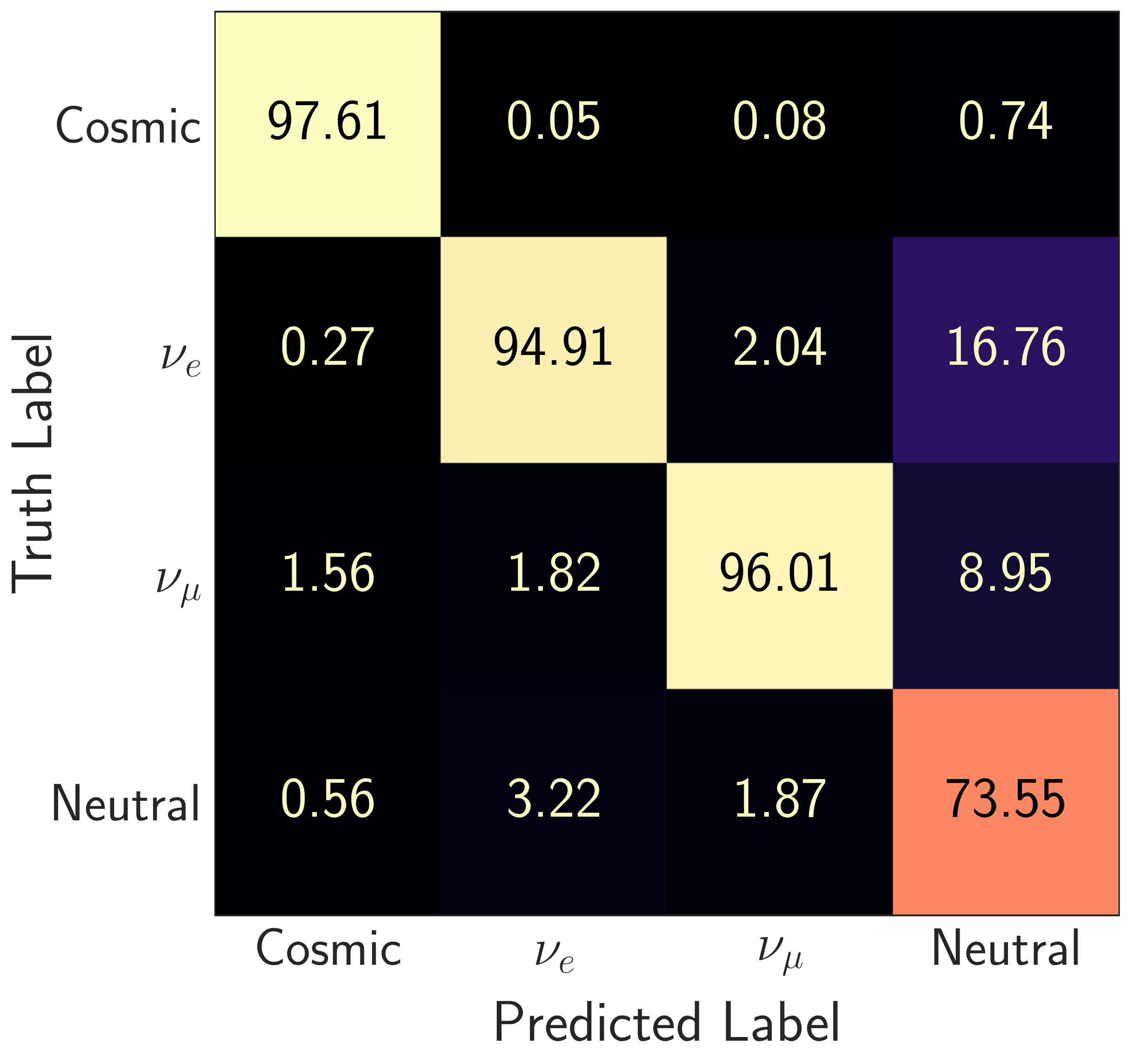}
  \caption{\textbf{Purity} matrix, normalized along predictions.}
  \label{figevent-confusion-purity-baseline-appendix}
\end{subfigure}
\hfill
\caption{EventCVN Baseline 4 Class Event Confusion Matrices.}
\label{fig:event-confusion-baseline-appendix}
\end{figure}

\begin{figure}[h]
\centering
\begin{subfigure}{.4\textwidth}
  \centering
  \includegraphics[width=1.0\linewidth]{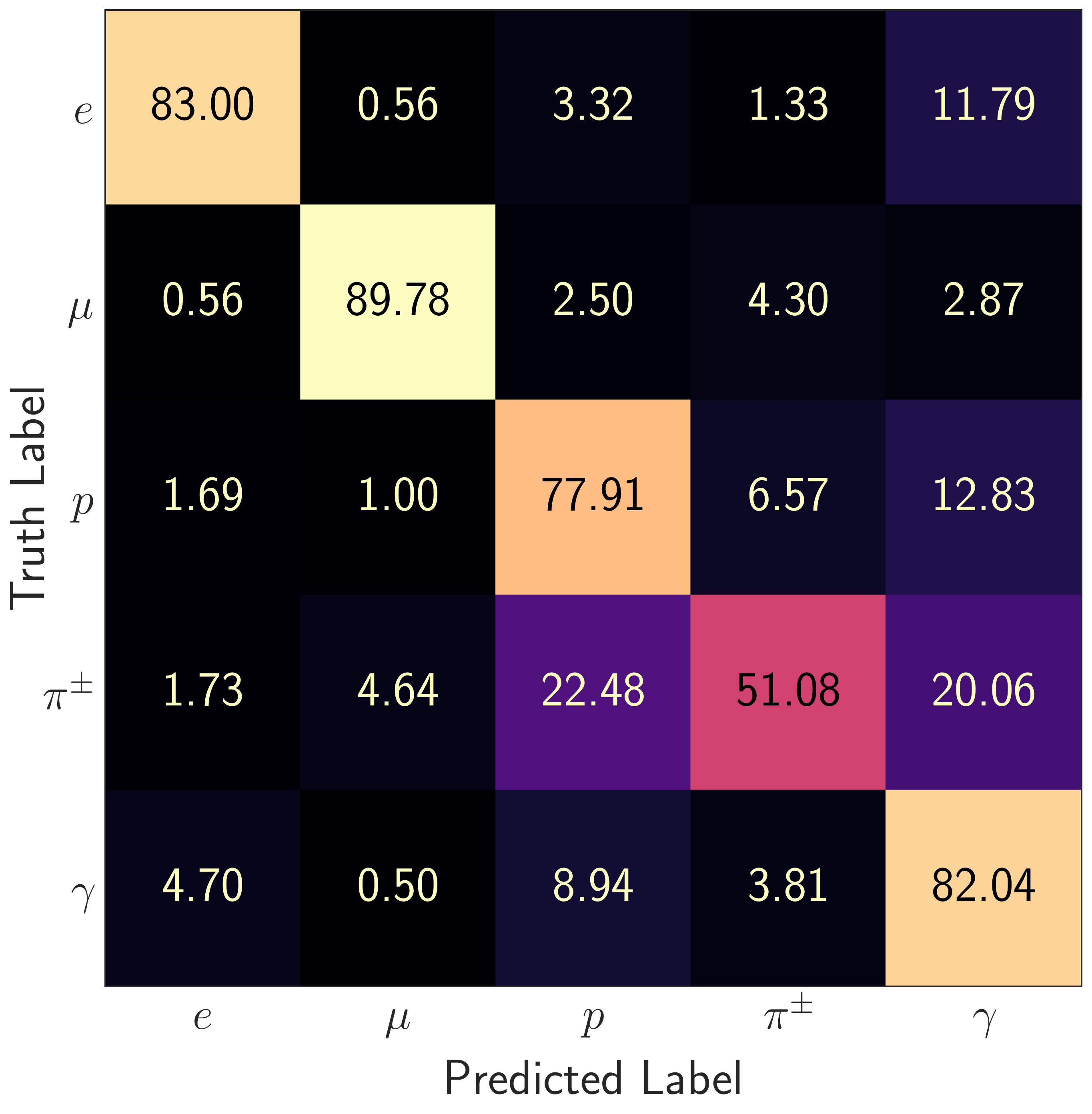}
  \caption{\textbf{Efficiency} matrix, normalized along truth labels.}
  \label{fig:prong_confusion_efficiency}
\end{subfigure}%
\hfill
\begin{subfigure}{.4\textwidth}
  \centering
  \includegraphics[width=1.0\linewidth]{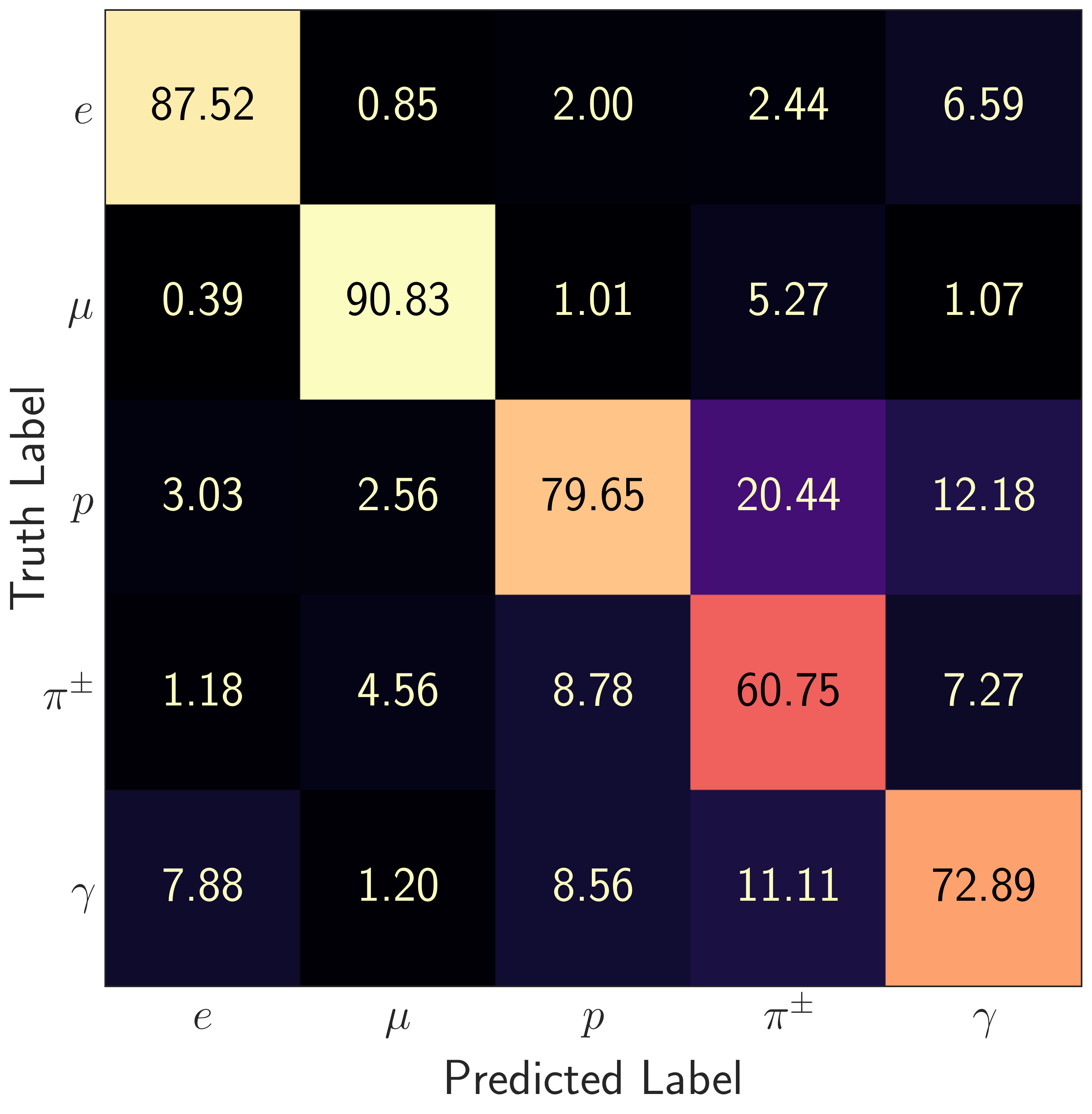}
  \caption{\textbf{Purity} matrix, normalized along predictions.}
  \label{fig:prong_confusion_purity}
\end{subfigure}
\hfill
    \caption{TransformerCVN 5-class Prong Confusion Matrices.}
\label{fig:prong}
\end{figure}

\begin{figure}[h]
\centering
\begin{subfigure}{.4\textwidth}
  \centering
  \includegraphics[width=1.0\linewidth]{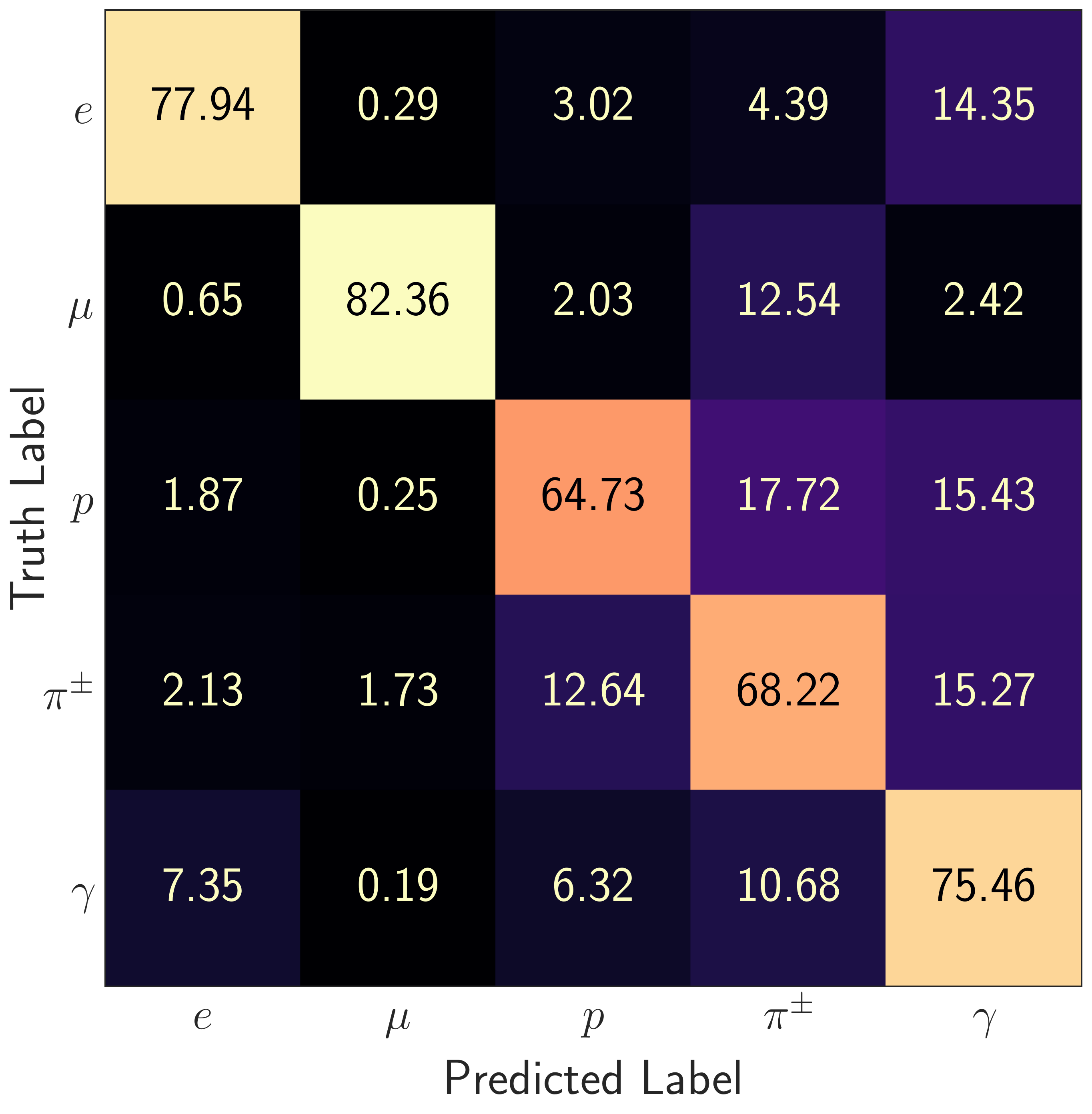}
  \caption{\textbf{Efficiency} matrix, normalized along truth labels.}
  \label{fig:event_confusion_efficiency}
\end{subfigure}%
\hfill
\begin{subfigure}{.4\textwidth}
  \centering
  \includegraphics[width=1.0\linewidth]{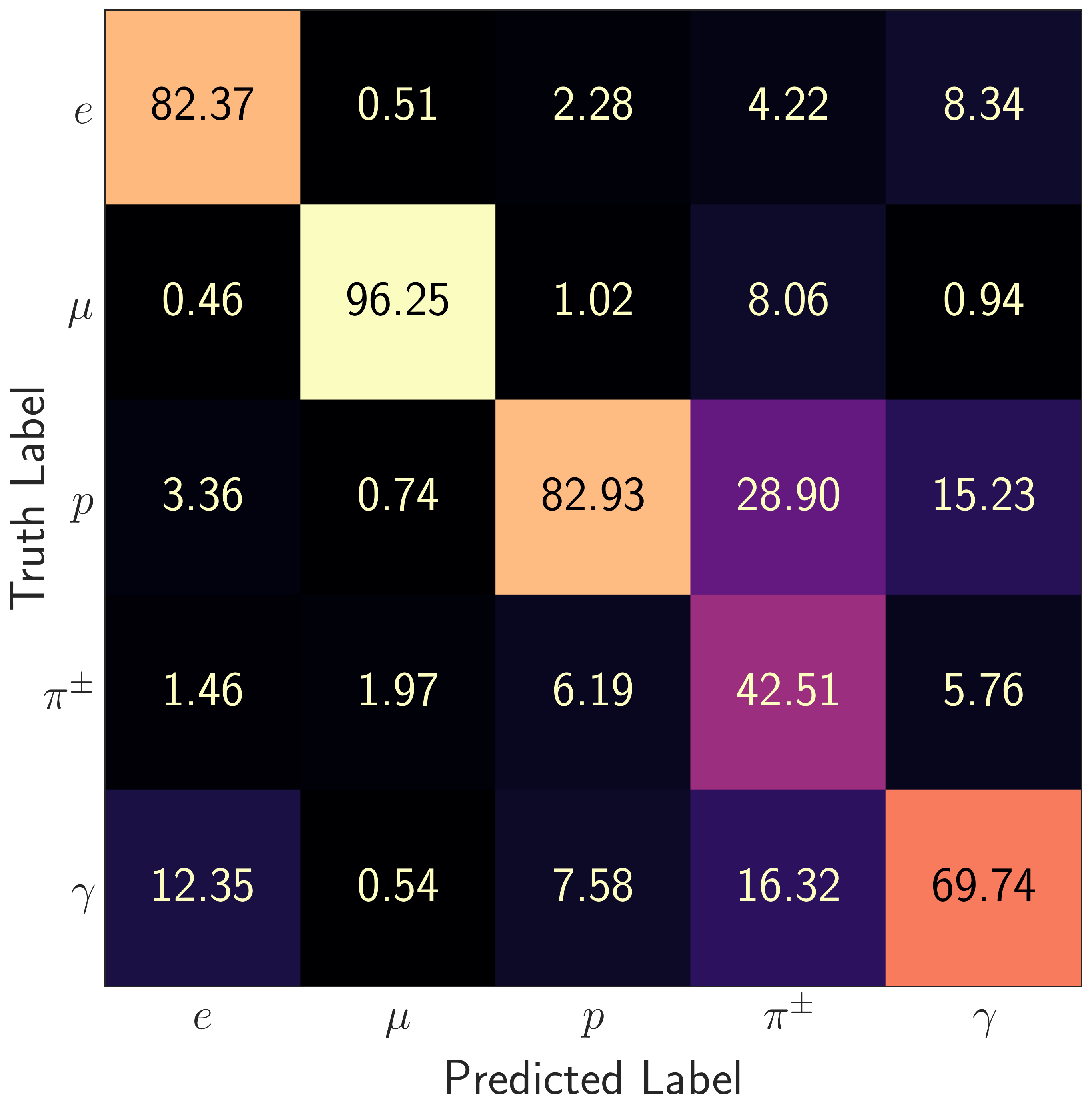}
  \caption{\textbf{Purity} matrix, normalized along predictions.}
  \label{fig:event_confusion_purity}
\end{subfigure}
\hfill
\caption{ProngCVN Baseline 5-class Prong Confusion Matrices.}
\label{fig:event_confusion}
\end{figure}

\begin{figure}[h]
\centering
\begin{subfigure}{.4\textwidth}
  \centering
  \includegraphics[width=1.0\linewidth]{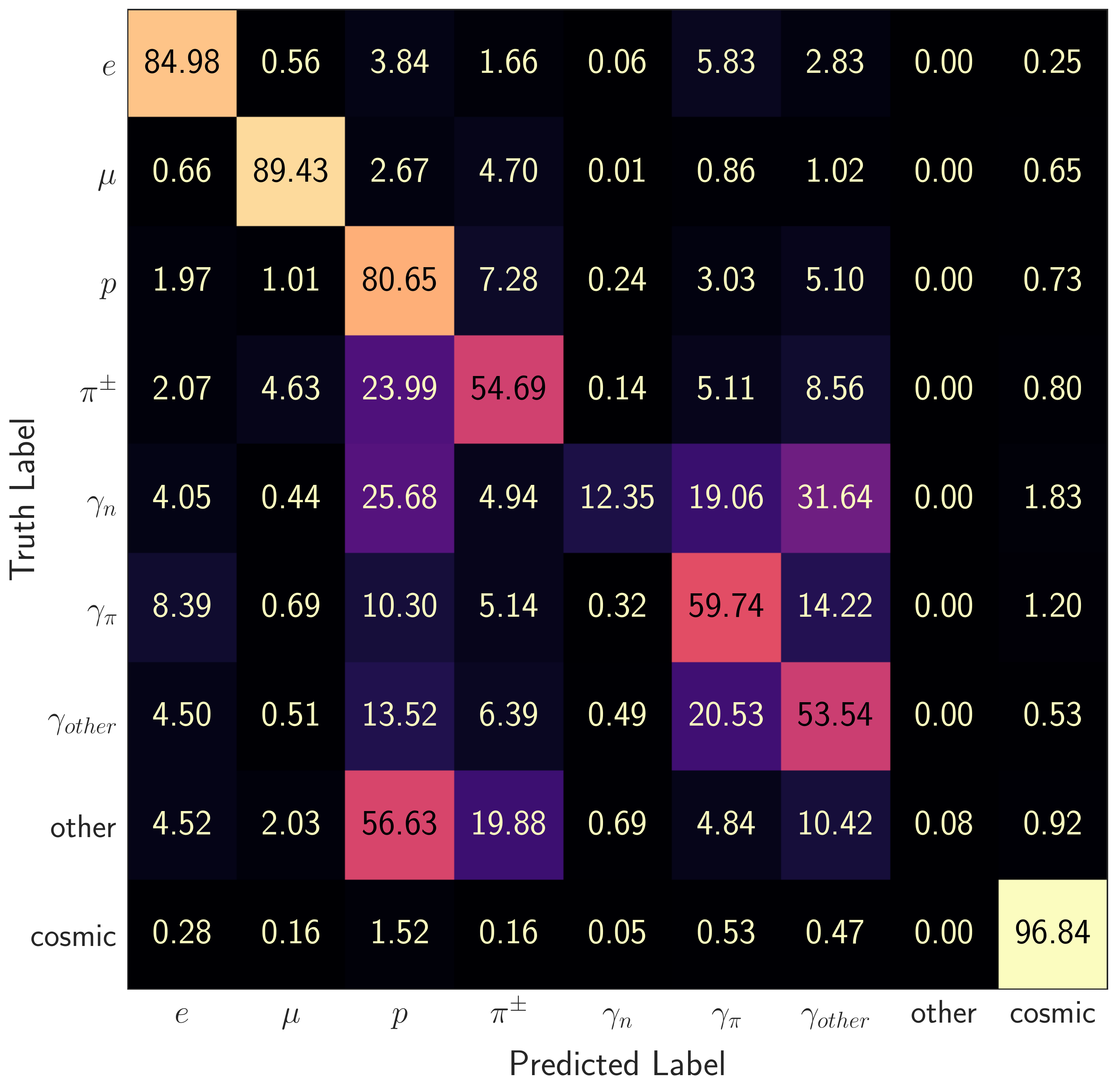}
  \caption{\textbf{Efficiency} matrix, normalized along truth labels.}
  \label{fig:prong_confusion_efficiency}
\end{subfigure}%
\hfill
\begin{subfigure}{.4\textwidth}
  \centering
  \includegraphics[width=1.0\linewidth]{Figures/Confusion/nova_prong_targets_confusion_pred.pdf}
  \caption{\textbf{Purity} matrix, normalized along predictions.}
  \label{fig:prong_confusion_purity}
\end{subfigure}
\hfill
\caption{TransformerCVN Full 9-class Prong Confusion Matrices.}
\label{fig:prong}
\end{figure}

\begin{figure}[h]
\centering
\begin{subfigure}{.48\textwidth}
  \centering
  \includegraphics[width=1.0\linewidth]{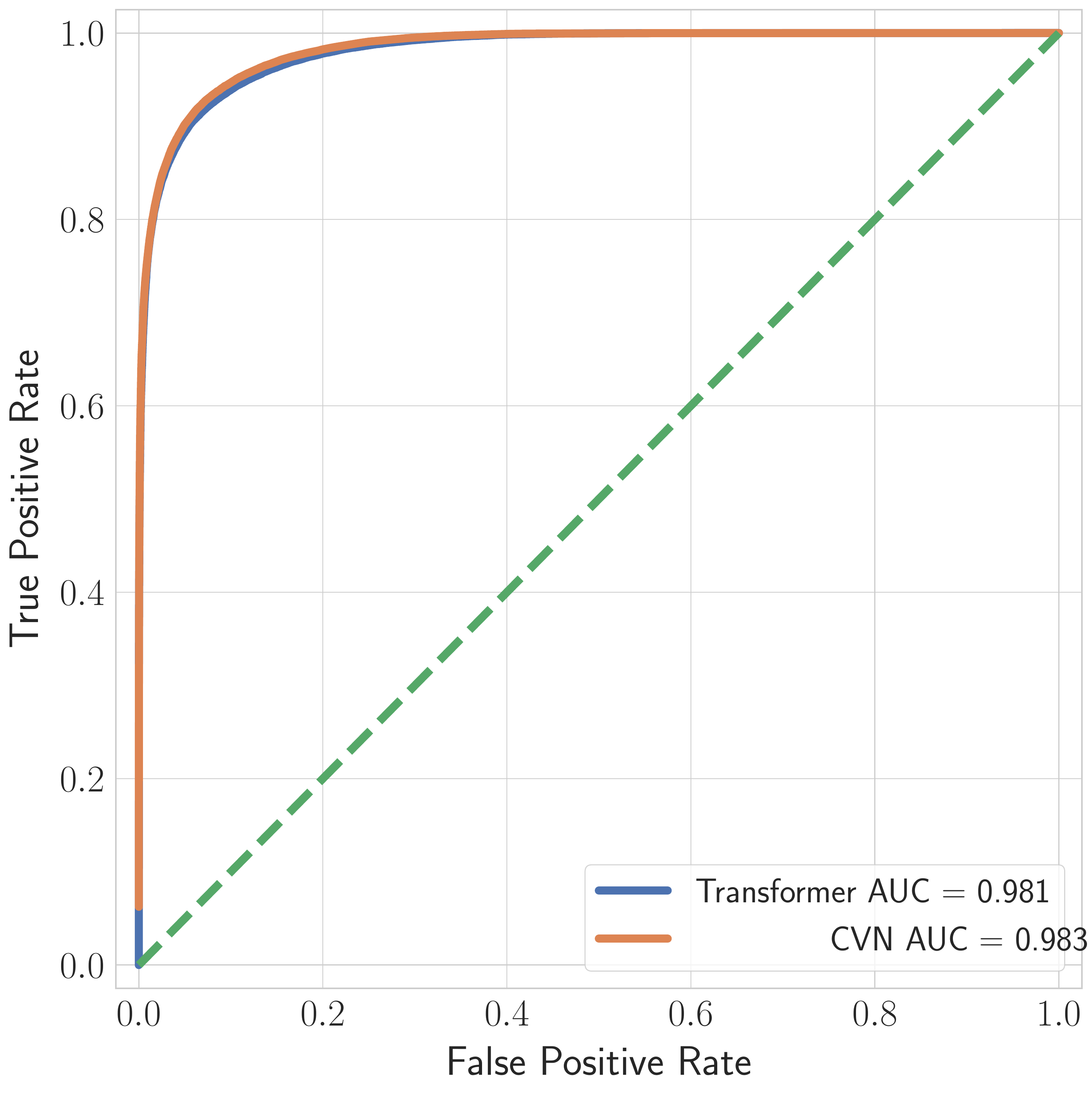}
  \caption{ROC Curve for $\nu_e$ event reconstruction.}
\end{subfigure}%
\hfill
\begin{subfigure}{.48\textwidth}
  \centering
  \includegraphics[width=1.0\linewidth]{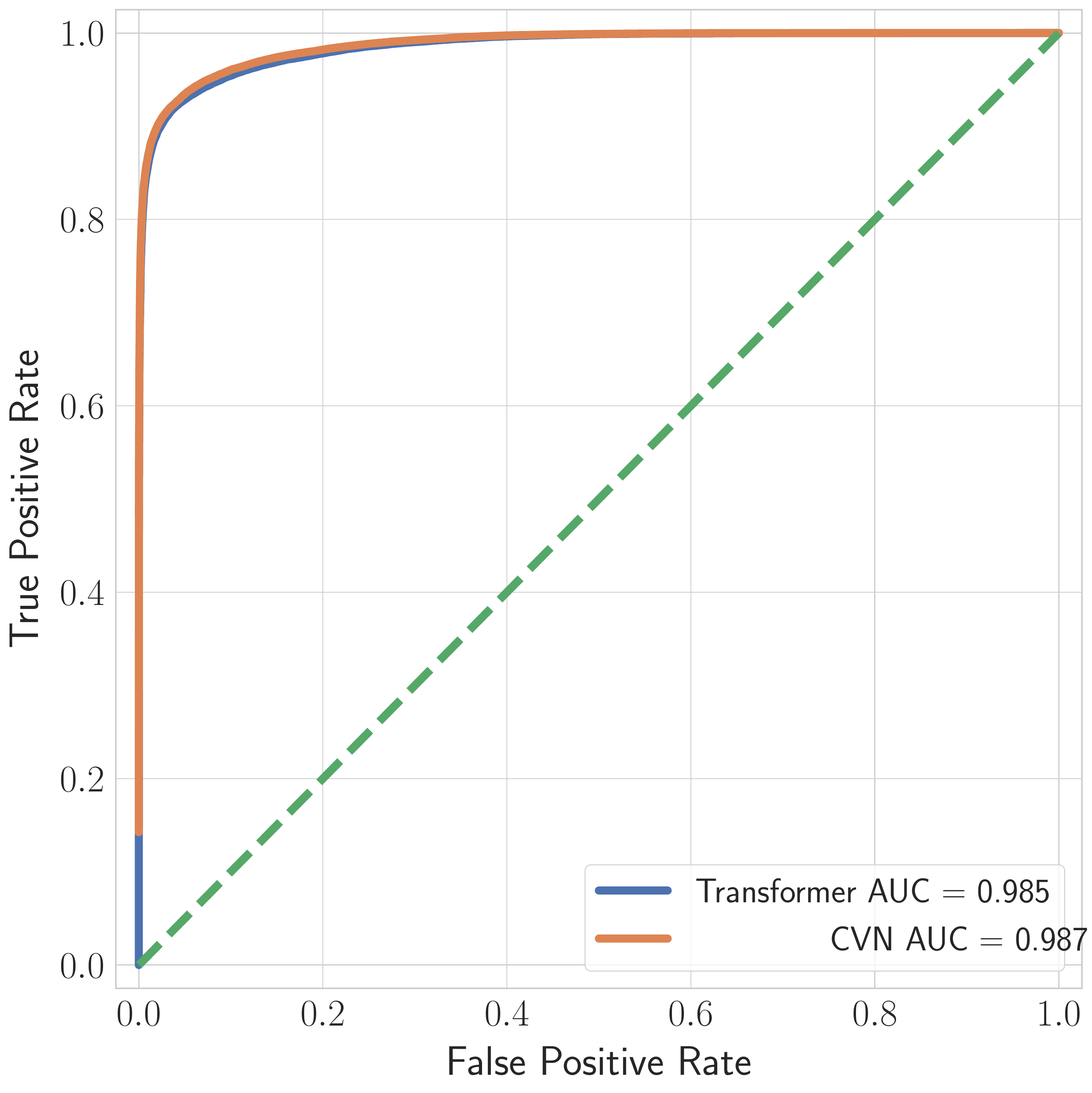}
  \caption{ROC Curve for $\nu_\mu$ event reconstruction.}
\end{subfigure}
\caption{Event classification ROC curves for TransformerCVN and EventCVN Baseline.}
\label{fig:event_roc_curves}
\end{figure}

\begin{figure}[h]
\centering
\begin{subfigure}{.48\textwidth}
  \centering
  \includegraphics[width=1.0\linewidth]{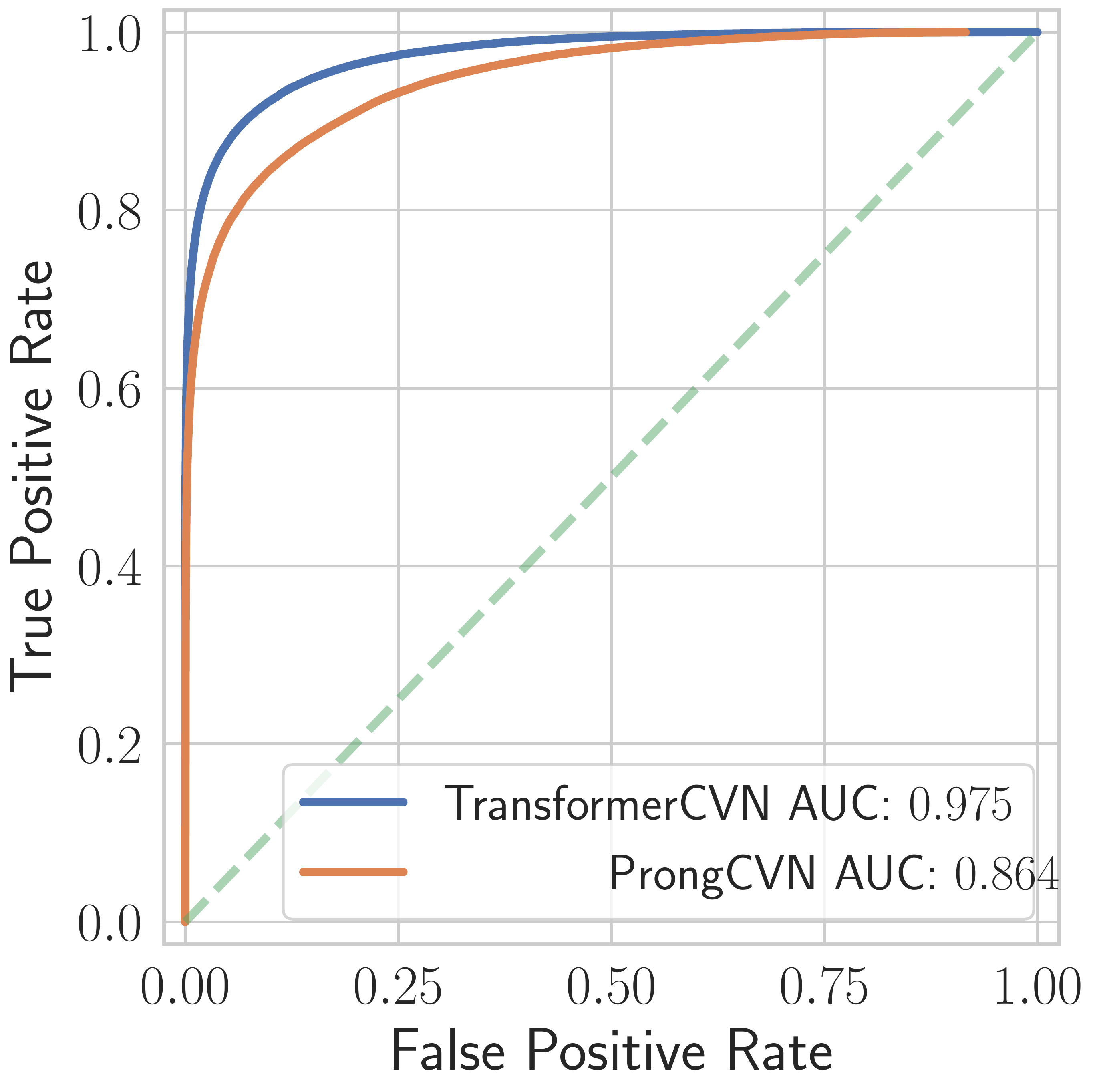}
  \caption{ROC Curve for $e$ prong reconstruction.}
\end{subfigure}%
\hfill
\begin{subfigure}{.48\textwidth}
  \centering
  \includegraphics[width=1.0\linewidth]{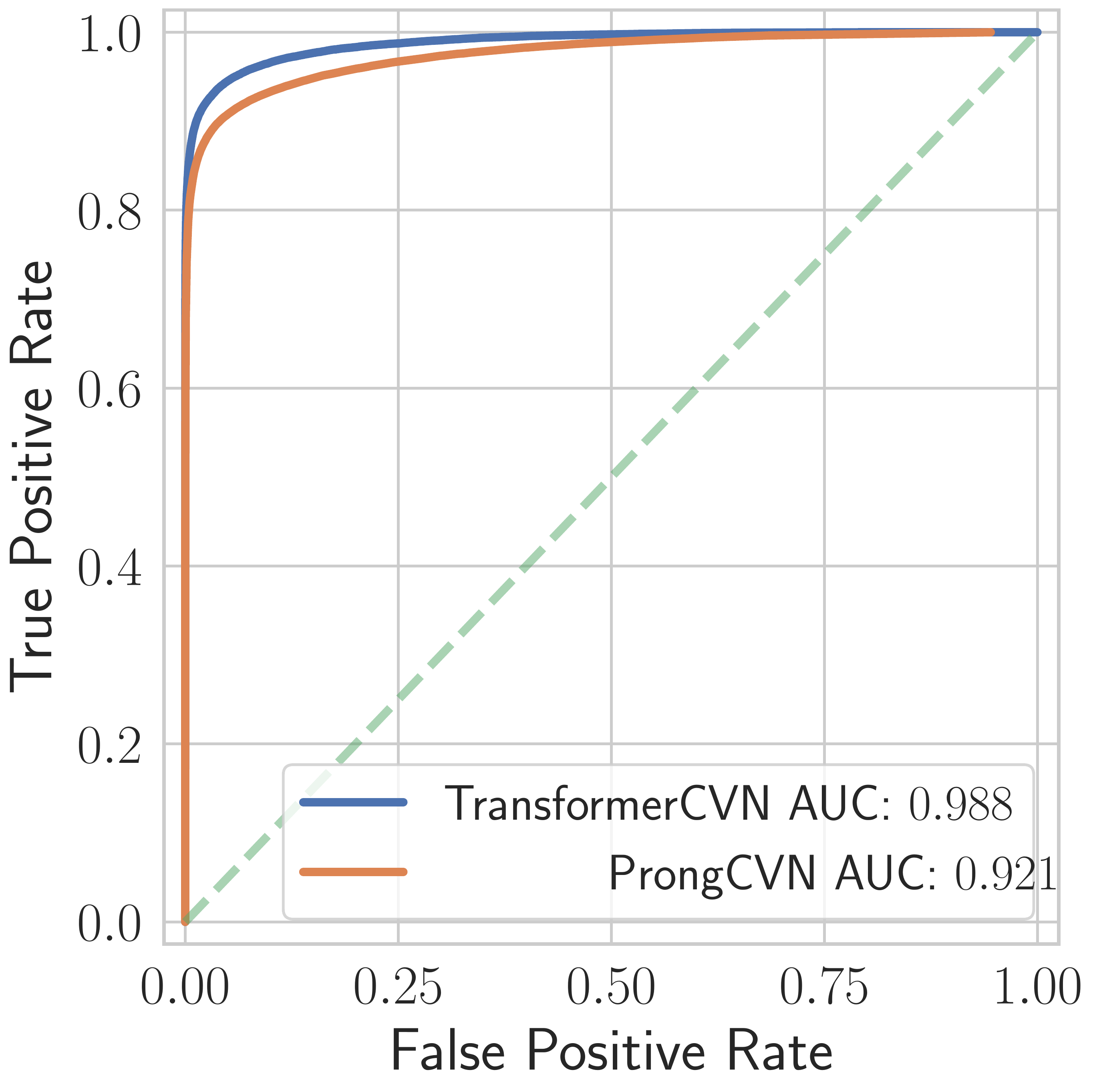}
  \caption{ROC Curve for $\mu$ prong reconstruction.}
\end{subfigure}
\caption{Lepton Prong Reconstruction ROC curves for TransformerCVN and ProngCVN Baseline.}
\label{fig:prong_roc_curves}
\end{figure}

\begin{figure}[h]
\centering
\begin{subfigure}{.48\textwidth}
  \centering
  \includegraphics[width=1.0\linewidth]{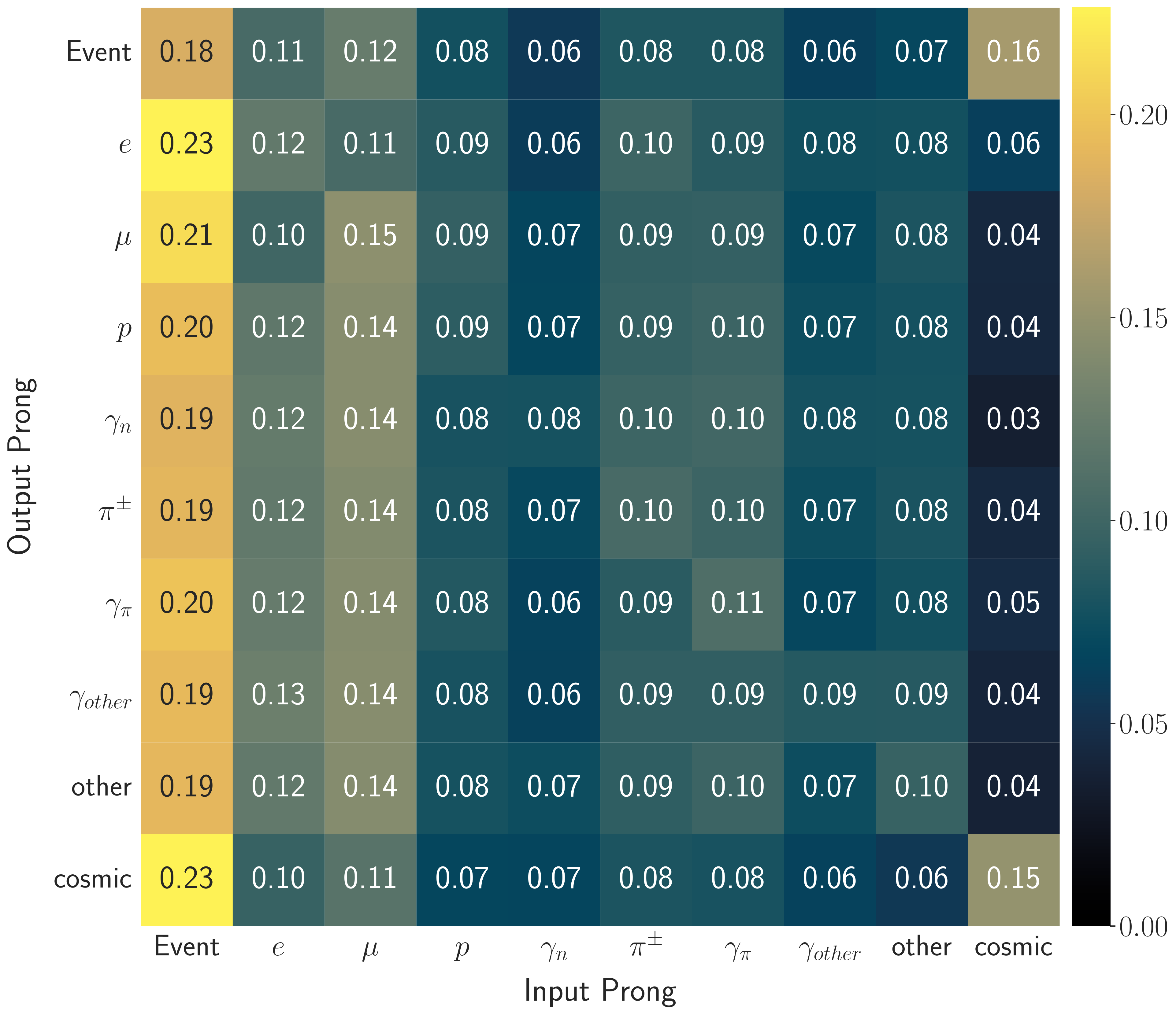}
  \caption{Average pairwise \textbf{Per-Prong} attention scores.}
  \vfill
\end{subfigure}%
\hfill
\begin{subfigure}{.48\textwidth}
  \centering
  \includegraphics[width=1.0\linewidth]{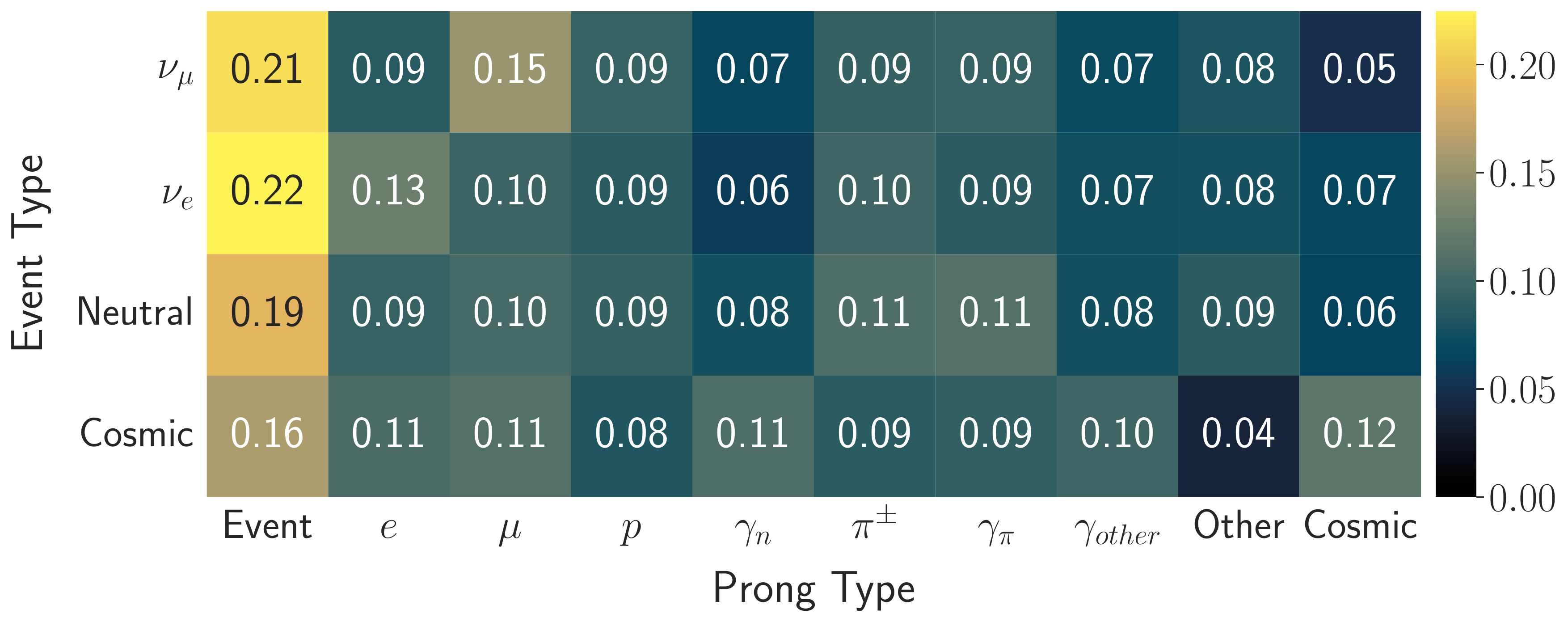}
  \caption{Average event \textbf{Per-Prong} attention scores.}
  \vfill
\end{subfigure}%
\caption{An Alternative aggregation method for attention matrices which just look at how each individual prong contributes to the attention score of prongs and events. This includes a downside in that prongs which appear more than once in events, such as photons, will be underrepresented since their attention will be split among other prongs with the same type. However, this does provide a more low-level view of the importance of different prongs in different classification and reconstruction tasks.}
\label{fig:prong_attention}
\end{figure}

\begin{figure}[h]
\centering
\begin{subfigure}{0.5\textwidth}
  \centering
  \includegraphics[width=1.0\linewidth]{Figures/normalized_projection_gradients_normalize_all.pdf}
  \caption{Flat weighted integrated salience}
  \label{fig:projected-saliance-flat}
\end{subfigure}\\
\begin{subfigure}{0.5\textwidth}
  \centering
  \includegraphics[width=1.0\linewidth]{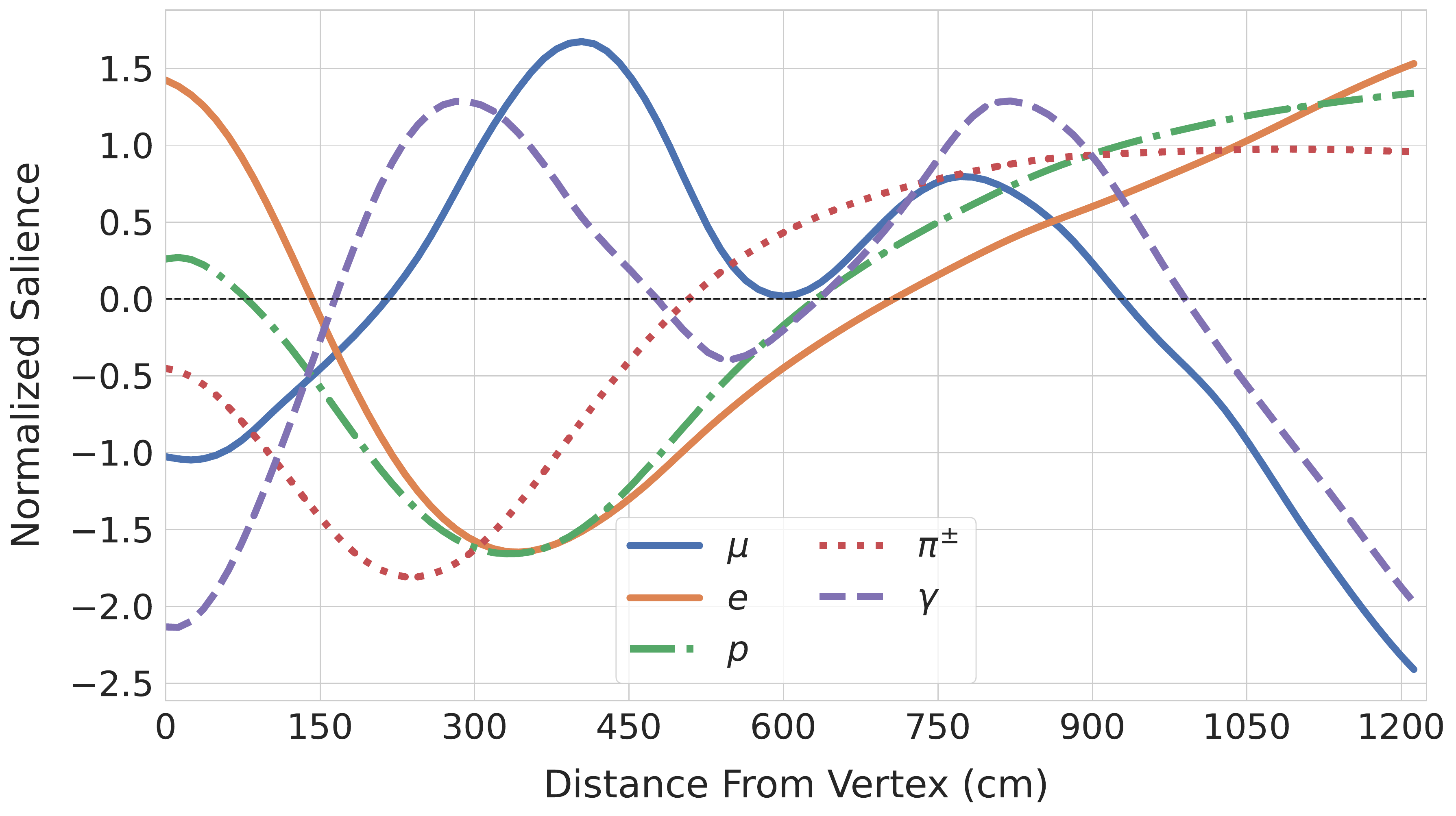}
  \caption{Gaussian weighted integrated salience.}
  \label{fig:projected-gaussian-saliance-flat}
\end{subfigure}
\caption{Full length views of the integrated salience maps along particle tracks which include the outer regions of the track. We additionally include a Gaussian-Weighted variant which puts more importance to salience near the particle's track. We re-weighting the salience maps with a Gaussian weight with respect to distance from the track's center before integrating the maps along their width.}
\label{fig:integrated-saliency}
\end{figure}

\end{document}